\pdfoutput=1

\documentclass[11pt]{article}

\usepackage{ACL2023}
\usepackage{times}
\usepackage{latexsym}

\usepackage[T1]{fontenc}

\usepackage[utf8]{inputenc}

\usepackage{microtype}

\usepackage{graphicx}
\usepackage{multicol}
\usepackage{multirow}
\usepackage{amsmath}
\usepackage{amssymb}
\usepackage{bbding}
\usepackage{subfigure}
\usepackage{colortbl}
\usepackage{pifont}
\usepackage{bbm}
\usepackage{makecell}
\usepackage{bbding}
\usepackage{mathrsfs}
\usepackage{bm}
\usepackage{CJKutf8}
\usepackage{booktabs}
\usepackage{tabu}
\usepackage{fontawesome}
\usepackage{mdframed}
\usepackage[linesnumbered,ruled,vlined]{algorithm2e}

\NewDocumentCommand{\zhangwei}
{ mO{} }{\textcolor{blue}{\textsuperscript{\textit{zhangwei}}\textsf{\textbf{\small[#1]}}}}

\newcommand{\instruct}{CodeSimpleQA-Instruct}
\newcommand{\sftmodel}{CodeSimpleQA-RFT}
\newcommand{\rlmodel}{CodeSimpleQA-RL}
\newcommand{\benchmark}{CodeSimpleQA}

\usepackage[most,skins,theorems]{tcolorbox}
\tcbuselibrary{skins,breakable,fitting,hooks}  
\tcbset{
  aibox/.style={
    width=\linewidth,
    top=7pt,
    bottom=2pt,
    colback=rliablered!18!white,
    colframe=black,
    colbacktitle=black,
    enhanced,
    center,
    attach boxed title to top left={yshift=-0.1in,xshift=0.15in},
    boxed title style={boxrule=0pt,colframe=white,},
  }
}
\tcbset{
  aibox2/.style={
    width=\linewidth,
    top=7pt,
    bottom=2pt,
    colback=green!18!white,
    colframe=black,
    colbacktitle=black,
    enhanced,
    center,
    attach boxed title to top left={yshift=-0.1in,xshift=0.15in},
    boxed title style={boxrule=0pt,colframe=white,},
  }
}
\newtcolorbox{AIbox}[2][]{aibox,title=#2,#1}
\newtcolorbox{AIbox2}[2][]{aibox2,title=#2,#1}

\title{\benchmark{}: Scaling Factuality in Code Large Language Models}

\author{
  Jian Yang\textsuperscript{\rm 1},
  {\bf Wei Zhang}\textsuperscript{\rm 1},
  {\bf Yizhi Li}\textsuperscript{\rm 2},
  {\bf Shawn Guo}\textsuperscript{\rm 1}, 
  {\bf Haowen Wang}\textsuperscript{\rm 1}, 
  {\bf Aishan Liu}\textsuperscript{\rm 1}, \\
  {\bf Ge Zhang}\textsuperscript{\rm 3}, 
  {\bf Zili Wang}\textsuperscript{\rm 4},
  {\bf Zhoujun Li}\textsuperscript{\rm 1}, 
  {\bf Xianglong Liu}\textsuperscript{\rm 1\thanks{\ \ Corresponding Author.}},
  {\bf Weifeng Lv}\textsuperscript{\rm 1*} \\
   \textsuperscript{\rm 1}Beihang University;
   \textsuperscript{\rm 2}Manchester;
   \textsuperscript{\rm 3}M-A-P; 
   \textsuperscript{\rm 4}StepFun; \\
   \texttt{\{jiayang\}@buaa.edu.cn} \\
}
\newmdenv[
  backgroundcolor=red!05,
  linecolor=quoteborder,
  skipabove=1em,
  skipbelow=0em,
  leftline=true,
  topline=false,
  bottomline=false,
  rightline=false,
  linecolor=red!66,
  linewidth=4pt
]{githubquote}
\begin{document}
\begin{CJK*}{UTF8}{gbsn}
\maketitle
\begin{abstract}
Large language models (LLMs) have made significant strides in code generation, achieving impressive capabilities in synthesizing code snippets from natural language instructions. However, a critical challenge remains in ensuring LLMs generate factually accurate responses about programming concepts, technical implementations, etc. Most previous code-related benchmarks focus on code execution correctness, overlooking the factual accuracy of programming knowledge. To address this gap, we present \benchmark{}, a comprehensive bilingual benchmark designed to evaluate the factual accuracy of code LLMs in answering code-related questions, which contains carefully curated question-answer pairs in both English and Chinese, covering diverse programming languages and major computer science domains. Further, we create \instruct{}, a large-scale instruction corpus with 66M samples, and develop a post-training framework combining supervised fine-tuning and reinforcement learning. Our comprehensive evaluation of diverse LLMs reveals that even frontier LLMs struggle with code factuality. Our proposed framework demonstrates substantial improvements over the base model, underscoring the critical importance of factuality-aware alignment in developing reliable code LLMs.
\end{abstract}

\section{Introduction}
Code large language models~\cite{qwen25coder,deepseek_coder,code_llama,claude4,gpt5} have made significant strides in code generation, achieving impressive capabilities in synthesizing code snippets and even generating complete functions from natural language instructions.
Meanwhile, a significant challenge in AI development is to ensure language models generate factually accurate responses about programming concepts, software engineering practices, and technical implementations.

\begin{figure}[t!]
\begin{center}
    \includegraphics[width=0.75\columnwidth]{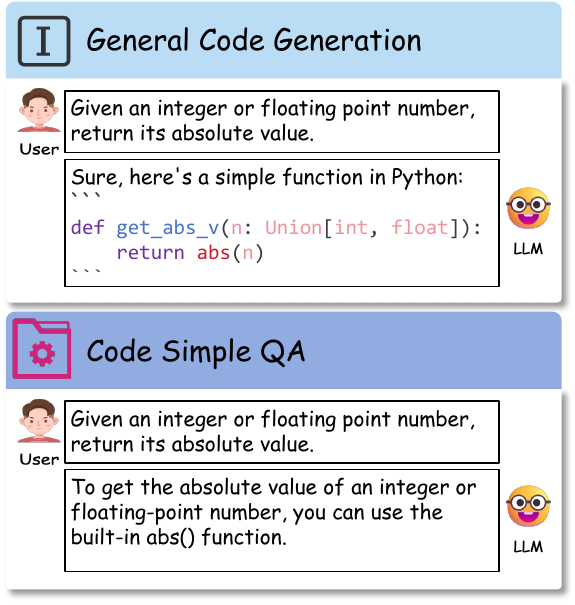}
    \vspace{-10pt}
    \caption{Comparison between the code generation task and the code factual question answering task.}
    \label{fig:intro}
    \vspace{-30pt}
\end{center}
\end{figure}
LLM factuality is the ability of LLMs to produce information verifiable by authoritative sources, evaluated using benchmarks that measure performance across short-form, long-form, and multilingual tasks. In \autoref{fig:intro}, the community has been focusing on code generation tasks, neglecting code question answering tasks.
In practice, many LLMs sometimes produce false answers, where factual inaccuracies can lead to bugs, security vulnerabilities, or inefficient implementations.
While existing benchmarks like SimpleQA~\cite{simpleqa} and Chinese SimpleQA~\citep{c_simpleqa} address the evaluation of the factual question answering (QA) in the general domain, there remains a gap in evaluating code-related factual knowledge of LLMs, specifically in software development contexts. 

\begin{figure*}[t]
\centering
\includegraphics[width=0.8\textwidth]{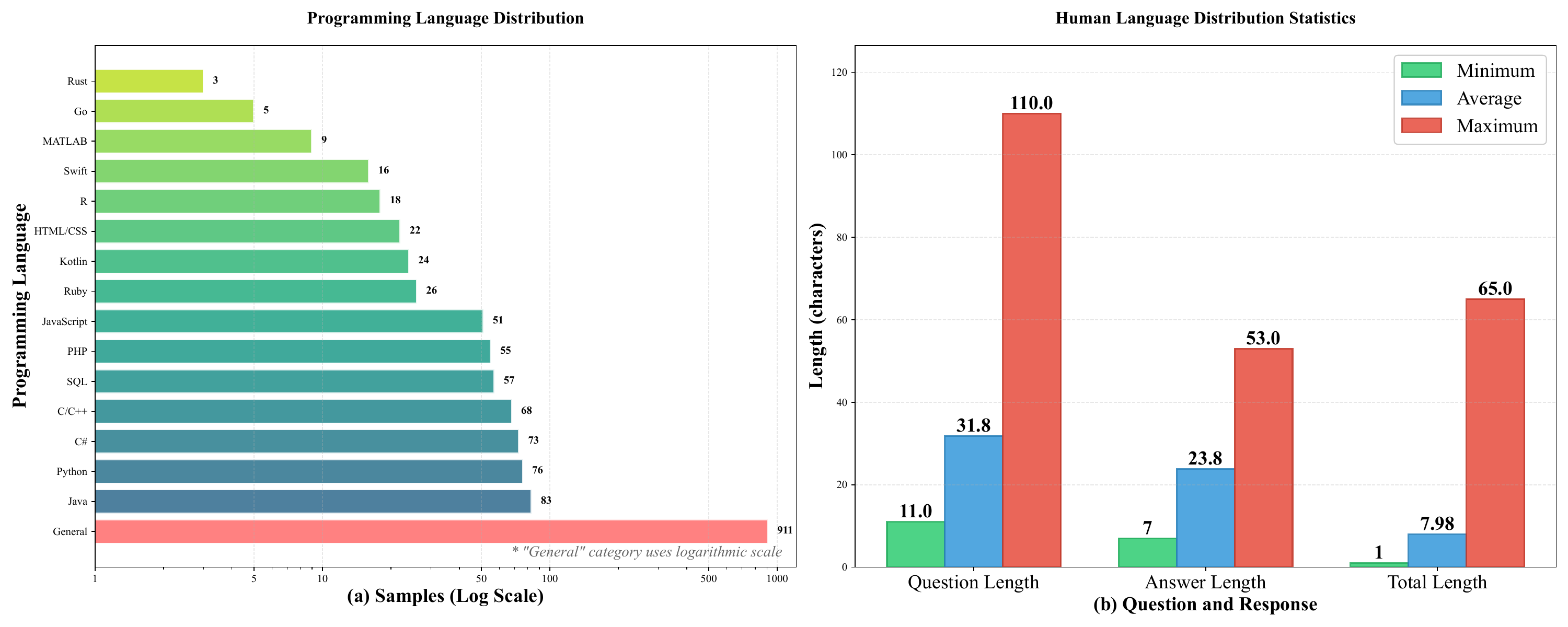}
\vspace{-10pt}
\caption{Data statistics for \benchmark{}.}
\label{fig:language_distribution}
\vspace{-20pt}
\end{figure*}
In this paper, we introduce \benchmark{}, a bilingual benchmark designed to evaluate the factual accuracy of code LLMs in answering code-related questions. The benchmark contains 1,498 carefully curated question-answer pairs (782 in English and 716 in Chinese) covering multiple programming languages and 8 major computer science domains, with all answers grounded in official documentation and verified through human expert annotation. To enable model improvement, we create \instruct{}, a massive training dataset of 66.9M samples, and trained models using SFT and GRPO. Their evaluation of numerous LLMs reveals that even frontier LLMs struggle with code factuality. \rlmodel{} demonstrates significant improvements over its base model, validating the effectiveness of its post-training approach for enhancing factual code knowledge in LLMs.

The contributions of this work are: (1) 
\textbf{\benchmark{}}: A comprehensive bilingual benchmark containing 1,498 human-curated factual question-answer pairs covering 15+ programming languages and 21 computer science domains, with answers grounded in code-related documents and verified by expert software engineers. (2) \textbf{\instruct{}}: A large-scale instruction-following dataset with 66.9M samples built through a systematic pipeline involving document recall, knowledge clustering, QA generation, and LLM-as-a-Judge verification. (3) \textbf{Training Methodology and Analysis}: A post-training framework combining SFT and GRPO that significantly improves factual code knowledge in LLMs. Our evaluation reveals: (a) RAG excels on frequently-updated documentation while SFT performs better on stable concepts; (b) test-time scaling shows rapid initial performance gains before plateauing; (c) thinking and non-thinking modes follow logarithmic scaling patterns with model parameters; and (d) domain-specific analysis shows models struggle with specialized areas like bioinformatics but perform better on web technologies and software engineering.

\section{\benchmark{}}

\paragraph{Dataset Statistics}
Our framework comprises instruction corpus \instruct{} with 67M samples (53.6M English, 13.4M Chinese) for post-training and test set \benchmark{} with 1,478 samples (762 English, 716 Chinese). As shown in \autoref{fig:language_distribution}, the language distribution is dominated by general questions (911 samples), followed by Java (83), Python (76), C\# (73), C/C++ (68), SQL (57), PHP (55), and JavaScript (51), while character length statistics for questions, answers, and totals are detailed in \autoref{tab:data_statistics}.

\begin{table}[t!]
\centering
\resizebox{\columnwidth}{!}{
\begin{tabular}{lrlr}
\toprule
\textbf{Statistics} & \textbf{Number} & \textbf{Statistics} & \textbf{Number} \\
\midrule
\textbf{\benchmark{}} & \multicolumn{3}{r}{$782/716$ (En/Zh)} \\
- Databases & $93$ & - Mobile Computing & $87$ \\
-Networks and Communications & $48$ & - Computer Systems & $27$ \\
-Software Engineering & $180$ & - Operating Systems & $125$ \\
-Web Technologies & $277$ & - Theory of Computation & $11$ \\
-Data Science and Analytics & $53$ & - Human-Computer Interaction & $31$ \\
- Programming Languages & $252$ & - Machine Learning & $21$ \\
- Computer Graphics & $47$ & - Emerging Technologies & $19$ \\
- Cybersecurity & $63$ & - Bioinformatics & $12$ \\
- Game Development & $59$ & - Algorithms and Data Structures & $17$ \\
- Information Systems & $28$ & - Artificial Intelligence & $25$ \\
- Computational Science & $23$ & & \\
\arrayrulecolor{black}\midrule
\textbf{\instruct{}} & \multicolumn{3}{r}{$53571094/13359625$ (En/Zh)} \\
\bottomrule
\end{tabular}}
\vspace{-10pt}
\caption{\benchmark{} dataset statistics.}
\vspace{-20pt}
\label{tab:data_statistics}
\end{table}

\paragraph{Human Annotation}
\benchmark{} is curated by 8 annotators and 3 engineers. We collect statements from multi code documentation sources (e.g., Stack Overflow, GitHub) and convert them to QA pairs, retaining 312 high-quality samples from 1.5K candidates after verification by at least 3 reviewers. More details can be found in Appendix~\ref{appendix: Human Annotation}.

\paragraph{Comparison between \benchmark{} and Other Benchmarks}

\autoref{tab:benchmark_compare} shows that \benchmark{} is the only benchmark specifically designed for computer science evaluation rather than general knowledge, offering bilingual support (Chinese and English) with an exceptionally large dataset size. Its diverse data sources, combining Code WebQA, self-constructed questions, and human-written content, ensure comprehensive coverage of CS knowledge, while employing LLM-as-a-Judge evaluation for more nuanced assessment of model capabilities.

\begin{figure*}[t!]
\begin{center}
    \includegraphics[width=0.65\textwidth]{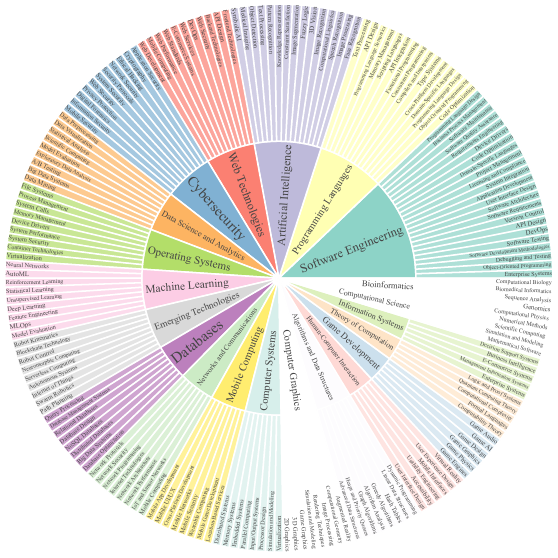}
    \vspace{-10pt}
    \caption{Examples of different domains in \benchmark{}.}
    \label{fig:model}
    \vspace{-15pt}
\end{center}
\end{figure*}

\begin{table*}[t!]
    \centering
    \small
    \resizebox{0.9\textwidth}{!}{
    \begin{tabular}{lcccccc}
        \toprule
         \textbf{Benchmark} & \textbf{Data Size} & \textbf{Language} & \textbf{Data Source} & \textbf{Domain} & \textbf{Instruct} & \textbf{Metric} \\
        \midrule
        WebQA~\citep{webqa} & 42187 & Chinese & Real World & General & \textcolor{red}{\ding{55}} & Accuracy \\
        MMLU~\citep{mmlu} & 15,908 & English & Exams \& Textbooks & General &\textcolor{red}{\ding{55}} & Accuracy \\
        CMMLU~\citep{cmmlu} & 11,528 & Chinese & Exams & General & \textcolor{red}{\ding{55}} & Accuracy \\
        MT-Bench~\citep{mt_bench} & 80 & English & Self-constructed & General &  \textcolor{red}{\ding{55}} & LLM-as-a-Judge \\
        C-Eval~\citep{ceval} & 13,948 & Chinese & Exams & General &  \textcolor{red}{\ding{55}} & Accuracy \\
        SimpleQA~\citep{simpleqa} & 4,326 & English & Human Writers & General  & \textcolor{red}{\ding{55}} & LLM-as-a-Judge \\
        Chinese SimpleQA~\citep{c_simpleqa} & 3000 & Chinese &   \makecell[c]{Self-constructed \\ Human Writers} & General & \textcolor{red}{\ding{55}} & LLM-as-a-Judge \\
        \midrule
        \textbf{\benchmark{} (ours)} & 1498/66930719 & Chinese/English & \makecell[c]{Code WebQA \\Self-constructed \\ Human Writers}  & Computer Science &  \textcolor{green}{\ding{51}} & LLM-as-a-Judge \\
        \bottomrule
    \end{tabular}}
    \vspace{-5pt}
    \caption{Comparisons between our Chinese SimpleQA and other benchmarks.}
    \label{tab:benchmark_compare}
    \vspace{-20pt}
\end{table*}

\section{Post-training for Factual Code QA}
\subsection{Dataset Construction.}
\paragraph{Recall the Code-related Document from Websites}
We propose a framework for constructing high-quality code knowledge QA pairs from Common Crawl by identifying code-related documents using fastText and applying multi-dimensional filtering. We prioritize source reliability by scoring domains (favoring official documentation, established technical blogs, Stack Overflow, and reputable tutorial platforms), assess content quality through LLM-based evaluation of technical depth and clarity, and verify appropriate code-text ratios to ensure documents contain both explanatory text and code examples. After extracting clean content via rule-based filtering to remove advertisements and HTML artifacts, we generate semantic embeddings using a specialized code-text BERT model and apply DBSCAN clustering to group documents into distinct knowledge clusters (e.g., API usage patterns, design principles, debugging techniques), from which we uniformly sample documents for QA pair generation.

\begin{figure*}[t]
\begin{center}
    \includegraphics[width=0.85\textwidth]{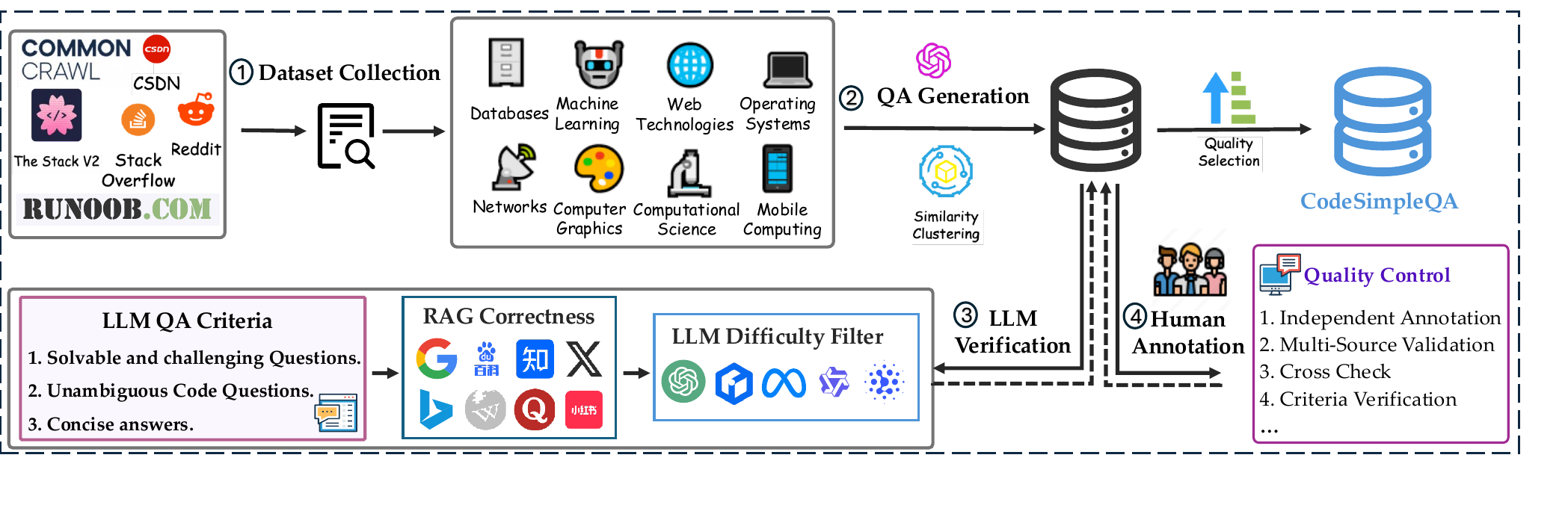}
    \vspace{-15pt}
    \caption{The framework to create the in \benchmark{}.}
    \label{fig:model}
    \vspace{-20pt}
\end{center}
\end{figure*}

\paragraph{Rewrite the Question and Answer Pair} LLMs are used to automatically generate factual QA pairs from each clustering (uniform sample) through a structured pipeline, incorporating explicit constraints ensuring objective questions with single, unambiguous, time-invariant answers, along with few-shot examples demonstrating correct formatting across categories. Deepseek-V3.1 generates the QA pair in the structured JSON outputs using low temperature settings ($0.1$) for consistency. This approach produces high-quality, objective technical assessments suitable for knowledge evaluation platforms while maintaining time-invariant accuracy and requiring minimal ongoing maintenance.

\paragraph{Correctness Verification}
To verify the rewritten question and answer pairs, we adopt the LLM-as-a-Judge~\cite{llm_as_a_judge_survey} to assess and score the quality of outputs based on the given original factual code documentation. We only keep the samples judged as the \textcolor{red}{CORRECT} label and fine-tune the base foundation model.
\begin{githubquote}
Question: \textcolor{blue}{\{question\}} \\
Predicted answer: \textcolor{blue}{\{predicted\_answer\}} \\
Factual Code Document: \textcolor{blue}{\{document\}} \\
Grade the predicted answer of this new question as one of: \\
\textcolor{red}{A: CORRECT.} \textcolor{red}{B: INCORRECT.}  \\
\textcolor{red}{C: NOT\_ATTEMPTED.} \\
Just return the letters 'A', 'B', or 'C', with no text around it.
\end{githubquote}

\subsection{Instruction Tuning}
\paragraph{Instruction Tuning}
We use the rejection-based supervised fine-tuning to train the base LLM to enhance knowledge about computer science. For factual question answering, we construct instruction-following datasets by formatting question and answer pair $(q,a)$ with explicit word count requirement for response. \textcolor{blue}{q=\{question\} + \textbackslash n + Response should not exceed 64 words}.

\subsection{Refinforcement Learning}
To enhance the capability of the code-related factual QA task, we adopt the group relative policy optimization (GRPO) with the help of the LLM-as-a-Judge. For a code-related question-answer pair $(q,a)$, we use the old poilcy $\pi_{\theta_{old}}$ to sample a group of $K$ individual responses $\{o_k\}_{k=1}^K$. Then, the advantage of the $i$-th response is calculated by:
\begin{equation}
\hat{A}_{k,t} = \frac{r_k - \text{mean}(\{r_k\}_{k=1}^K)}{\text{std}(\{r_k\}_{i=k}^K)}.
\end{equation}
where $r_{k}=\mathbb{I}(o;a)$, where $\mathbb{I}(\cdot)$ is the indicator function. $\mathbb{I}(o;a)=1$ when the model output $o$ by $\pi_{old}$ matches the groundtruth, else $\mathbb{I}(o;a)=0$.

Given the question and answer pair $(q,a)$, we optimize the the SFT model \instruct{} with the GRPO objective:
\begin{MiddleEquation}
\begin{equation}
\begin{aligned}
\mathcal{L}_\text{GRPO}(\theta)& = \mathbb{E}_{(q,a),\{o_k\}_{k=1}^K\sim \pi_{\theta_\text{old}}(\cdot\mid q)} \\&
\Bigg[ \frac{1}{K}\sum_{k=1}^{K} \frac{1}{|o_i|}\sum_{t=1}^{|o_k|} \Bigg( 
\min \Big( A_{1}, A_{2}) + \mathcal{L}_{KL}
\Bigg) \Bigg]
\label{eq:grpoloss}
\end{aligned}
\end{equation}
\end{MiddleEquation}where $A_{1} = r_{i,t}(\theta) \hat{A}_{i,t}$, $A_{2} = (r_{i,t}(\theta), 1 - \varepsilon, 1 + \varepsilon) \hat{A}_{i,t}$, and $\mathcal{L}_{KL} = - \beta D_{\text{KL}}(\pi_{\theta} || \pi_{\text{ref}})$. The importance sampling ratio is $r_{k,t}(\theta)=\frac{\pi_{\theta}(o_{k,t} \mid q, o_{k,<t})}{\pi_{\theta_{\text{old}}}(o_{k,t} \mid q,o_{k,<t})}$.

\begin{table*}[t!]
\centering
\resizebox{1.0\textwidth}{!}{
\begin{tabular}{l|c|ccccccccccccccccccccc|c}
\toprule
Model & Params & ADS & AI & BIO & CS & CG & CSYS & CYB & DSA & DB & ET & GD & HCI & IS & ML & MC & NC & OS & PL & SE & TOC & WT & Avg. \\
\midrule
\multicolumn{24}{c}{\textit{\textbf{Proprietary Large Language Models}}} \\ \midrule
gpt-4o-2024-11-20 & \faLock{} & 75.0 & 78.6 & 60.0 & 87.5 & 50.0 & 41.7 & 51.0 & 73.3 & 44.0 & 75.0 & 32.0 & 38.5 & 60.0 & 55.6 & 46.6 & 50.0 & 55.1 & 53.7 & 48.0 & 40.0 & 43.2 & 50.7 \\
chatgpt-4o-latest & \faLock{} & 50.0 & 73.7 & 46.2 & 70.0 & 63.3 & 54.5 & 51.4 & 78.9 & 51.4 & 42.9 & 43.8 & 52.9 & 46.2 & 53.8 & 49.0 & 50.0 & 59.2 & 49.4 & 45.1 & 55.6 & 42.0 & 51.4 \\
o3 & \faLock{} & 75.0 & 68.4 & 71.4 & 80.0 & 70.0 & 58.8 & 67.6 & 68.4 & 61.8 & 57.1 & 56.2 & 58.8 & 46.2 & 61.5 & 63.3 & 57.1 & 73.2 & 59.1 & 57.3 & 66.7 & 57.0 & 62.1 \\
o3-mini-2025-01-31 & \faLock{} & 50.0 & 68.4 & 42.9 & 100.0 & 66.7 & 47.1 & 45.9 & 73.7 & 51.4 & 85.7 & 34.9 & 47.1 & 53.8 & 53.8 & 61.9 & 50.0 & 57.7 & 47.2 & 47.9 & 47.1 & 52.3 & 53.4 \\
o4-mini & \faLock{} & 50.0 & 72.2 & 50.0 & 100.0 & 63.3 & 46.7 & 54.3 & 84.2 & 58.8 & 83.3 & 41.4 & 60.0 & 50.0 & 53.8 & 56.5 & 44.4 & 70.6 & 46.1 & 40.0 & 77.8 & 54.6 & 55.5 \\
claude-sonnet-4-20250514-thinking & \faLock{} & 75.0 & 73.3 & 57.1 & 100.0 & 53.8 & 47.6 & 58.1 & 80.0 & 54.2 & 44.4 & 51.2 & 31.6 & 36.4 & 70.0 & 59.8 & 60.0 & 64.5 & 46.2 & 50.9 & 76.9 & 47.1 & 55.4 \\
claude-sonnet-4-20250514 & \faLock{} & 66.7 & 61.5 & 28.6 & 85.7 & 60.0 & 51.9 & 51.9 & 53.3 & 54.5 & 36.4 & 38.1 & 33.3 & 33.3 & 42.9 & 58.3 & 70.3 & 64.3 & 54.7 & 44.1 & 76.9 & 49.4 & 53.5 \\
gpt-5-mini & \faLock{} & 40.0 & 78.6 & 100.0 & 85.7 & 58.5 & 50.0 & 58.3 & 69.6 & 57.8 & 100.0 & 44.4 & 71.0 & 55.6 & 66.7 & 68.6 & 46.7 & 66.7 & 45.8 & 52.7 & 0.0 & 54.0 & 57.8 \\
gpt-5 & \faLock{} & 75.0 & 90.9 & 80.0 & 100.0 & 77.8 & 62.5 & 75.5 & 84.6 & 60.0 & 50.0 & 57.1 & 50.0 & 66.7 & 55.6 & 74.2 & 62.1 & 88.4 & 58.1 & 67.4 & 80.0 & 47.5 & 67.2 \\
\midrule\multicolumn{24}{c}{\textit{\textbf{0.5B+ Large Language Models}}} \\ \midrule
Qwen2.5-0.5B-Instruct & 0.5B & 25.0 & 11.8 & 15.4 & 0.0 & 13.8 & 0.0 & 24.7 & 21.6 & 5.7 & 30.8 & 12.9 & 24.2 & 0.0 & 24.0 & 12.5 & 25.5 & 12.8 & 12.9 & 10.1 & 22.2 & 11.3 & 13.5 \\
Qwen2.5-Coder-0.5B-Instruct & 0.5B & 25.0 & 5.4 & 0.0 & 11.1 & 14.5 & 6.5 & 8.5 & 21.1 & 2.9 & 0.0 & 9.8 & 13.3 & 16.7 & 30.8 & 14.9 & 7.3 & 14.3 & 14.0 & 16.6 & 0.0 & 15.1 & 13.0 \\
Qwen3-0.6B & 0.6B & 0.0 & 15.8 & 18.2 & 44.4 & 10.2 & 18.2 & 16.2 & 21.1 & 5.7 & 0.0 & 9.7 & 17.6 & 16.7 & 15.4 & 18.9 & 18.5 & 8.6 & 14.0 & 7.5 & 22.2 & 14.5 & 13.4 \\
Llama-3.2-1B-Instruct & 1B & 0.0 & 15.8 & 14.3 & 0.0 & 6.7 & 6.1 & 5.4 & 27.0 & 5.7 & 14.3 & 9.7 & 12.1 & 32.0 & 24.0 & 10.2 & 14.5 & 11.6 & 12.4 & 12.6 & 11.1 & 10.2 & 11.5 \\
Qwen2.5-Coder-1.5B-Instruct & 1.5B & 50.0 & 16.2 & 28.6 & 21.1 & 16.9 & 5.9 & 10.8 & 27.0 & 17.1 & 14.3 & 9.4 & 18.2 & 23.1 & 30.8 & 16.5 & 25.0 & 25.5 & 19.2 & 12.7 & 11.1 & 16.5 & 17.9 \\
Qwen2.5-1.5B-Instruct & 1.5B & 85.7 & 26.3 & 14.3 & 40.0 & 23.7 & 12.1 & 21.6 & 47.4 & 14.5 & 28.6 & 6.3 & 24.2 & 15.4 & 53.8 & 20.4 & 21.8 & 21.1 & 16.9 & 13.8 & 22.2 & 16.3 & 20.0 \\
DeepSeek-R1-Distill-Qwen-1.5B & 1.5B & 75.0 & 16.7 & 0.0 & 31.6 & 6.9 & 6.2 & 5.8 & 17.6 & 3.0 & 14.3 & 3.6 & 11.8 & 7.7 & 30.8 & 2.3 & 10.9 & 0.0 & 3.8 & 2.6 & 11.8 & 3.2 & 6.2 \\
Llama-3.2-3B-Instruct & 3B & 25.0 & 21.1 & 14.3 & 10.0 & 16.9 & 11.8 & 10.8 & 15.8 & 14.3 & 14.3 & 12.7 & 18.2 & 30.8 & 30.8 & 14.4 & 10.9 & 15.5 & 19.3 & 12.7 & 22.2 & 16.0 & 15.8 \\
Qwen2.5-Coder-3B-Instruct & 3B & 50.0 & 31.6 & 42.9 & 11.1 & 27.1 & 11.8 & 30.1 & 36.8 & 23.5 & 28.6 & 9.7 & 36.4 & 23.1 & 61.5 & 28.9 & 25.0 & 25.9 & 22.6 & 19.0 & 35.3 & 25.6 & 25.5 \\
Qwen2.5-3B-Instruct & 3B & 50.0 & 16.2 & 14.3 & 40.0 & 13.6 & 12.1 & 24.7 & 36.8 & 14.3 & 14.3 & 9.4 & 42.4 & 24.0 & 46.2 & 14.4 & 32.7 & 17.0 & 20.7 & 16.2 & 23.5 & 18.1 & 20.0 \\
Qwen3-4B & 4B & 50.0 & 54.1 & 28.6 & 63.2 & 40.0 & 24.2 & 35.6 & 42.1 & 31.9 & 57.1 & 25.0 & 23.5 & 30.8 & 53.8 & 33.0 & 28.6 & 36.5 & 31.0 & 31.2 & 50.0 & 33.5 & 34.5 \\
Qwen3-4B-Thinking-2507 & 4B & 75.0 & 78.9 & 14.3 & 70.0 & 43.3 & 35.3 & 41.7 & 47.4 & 45.7 & 42.9 & 25.0 & 42.4 & 32.0 & 38.5 & 30.6 & 35.7 & 39.4 & 31.6 & 29.6 & 66.7 & 39.2 & 38.3 \\
\midrule\multicolumn{24}{c}{\textit{\textbf{7B+ Large Language Models}}} \\ \midrule
DeepSeek-R1-Distill-Qwen-7B & 7B & 50.0 & 31.6 & 14.3 & 30.0 & 13.3 & 11.8 & 16.4 & 31.6 & 8.6 & 0.0 & 9.5 & 12.1 & 8.0 & 32.0 & 6.2 & 7.3 & 10.1 & 12.6 & 5.0 & 44.4 & 5.0 & 11.6 \\
Qwen2.5-Coder-7B-Instruct & 7B & 50.0 & 31.6 & 28.6 & 40.0 & 30.5 & 17.6 & 16.4 & 36.8 & 20.3 & 42.9 & 15.6 & 41.2 & 15.4 & 23.1 & 26.8 & 25.0 & 34.5 & 28.2 & 25.9 & 11.1 & 31.5 & 27.6 \\
Qwen2.5-7B-Instruct & 7B & 75.0 & 48.6 & 28.6 & 42.1 & 23.3 & 17.6 & 27.0 & 36.8 & 22.9 & 42.9 & 15.6 & 30.3 & 15.4 & 46.2 & 27.1 & 36.4 & 28.4 & 24.0 & 20.9 & 22.2 & 22.3 & 26.3 \\
Qwen3-1.7B & 7B & 50.0 & 52.6 & 14.3 & 50.0 & 36.7 & 17.6 & 19.2 & 57.9 & 23.2 & 42.9 & 21.9 & 17.6 & 24.0 & 48.0 & 26.5 & 18.2 & 21.4 & 13.5 & 18.3 & 47.1 & 16.0 & 23.4 \\
Hunyuan-7B-Instruct & 7B & 50.0 & 31.6 & 28.6 & 40.0 & 26.7 & 12.1 & 29.7 & 21.6 & 14.3 & 14.3 & 19.0 & 11.8 & 23.1 & 38.5 & 14.4 & 25.0 & 23.0 & 25.0 & 13.7 & 11.1 & 18.3 & 21.0 \\
Llama-3.1-8B-Instruct & 8B & 25.0 & 21.1 & 28.6 & 70.0 & 26.7 & 24.2 & 18.9 & 42.1 & 22.9 & 57.1 & 12.5 & 36.4 & 30.8 & 38.5 & 26.8 & 28.6 & 32.4 & 27.0 & 13.6 & 33.3 & 24.1 & 26.0 \\
DeepSeek-R1-Distill-Llama-8B & 8B & 50.0 & 21.1 & 14.3 & 31.6 & 16.7 & 17.6 & 8.7 & 31.6 & 11.9 & 28.6 & 3.2 & 30.3 & 33.3 & 38.5 & 14.9 & 10.9 & 18.4 & 16.2 & 12.8 & 22.2 & 15.4 & 16.7 \\
Seed-Coder-8B-Instruct & 8B & 75.0 & 21.1 & 14.3 & 30.0 & 40.0 & 24.2 & 21.6 & 32.4 & 22.9 & 42.9 & 9.4 & 35.3 & 16.0 & 46.2 & 35.4 & 35.7 & 32.9 & 32.6 & 26.3 & 33.3 & 31.5 & 29.8 \\
Qwen3-8B & 8B & 57.1 & 57.9 & 46.2 & 52.6 & 37.3 & 29.4 & 43.2 & 57.9 & 45.7 & 85.7 & 28.6 & 29.4 & 15.4 & 38.5 & 31.2 & 35.7 & 44.0 & 33.0 & 30.9 & 47.1 & 29.7 & 36.8 \\
DeepSeek-R1-0528-Qwen3-8B & 8B & 75.0 & 32.4 & 30.8 & 21.1 & 38.5 & 24.2 & 36.1 & 48.5 & 26.9 & 50.0 & 32.3 & 44.4 & 33.3 & 24.0 & 29.2 & 34.6 & 32.1 & 27.2 & 21.3 & 35.3 & 24.6 & 30.0 \\
Qwen2.5-Coder-14B-Instruct & 14B & 50.0 & 47.4 & 42.9 & 31.6 & 30.0 & 42.4 & 29.7 & 36.8 & 28.6 & 57.1 & 19.0 & 41.2 & 23.1 & 46.2 & 38.8 & 46.4 & 42.3 & 31.5 & 32.1 & 33.3 & 30.2 & 34.5 \\
Qwen2.5-14B-Instruct & 14B & 75.0 & 47.4 & 28.6 & 50.0 & 37.9 & 29.4 & 37.8 & 52.6 & 43.5 & 42.9 & 12.9 & 41.2 & 15.4 & 30.8 & 31.2 & 32.7 & 38.0 & 33.9 & 23.2 & 55.6 & 23.1 & 32.5 \\
DeepSeek-R1-Distill-Qwen-14B & 14B & 50.0 & 48.6 & 28.6 & 50.0 & 30.0 & 32.3 & 30.1 & 42.1 & 31.9 & 42.9 & 9.4 & 24.2 & 15.4 & 61.5 & 39.6 & 25.5 & 35.7 & 32.6 & 25.8 & 35.3 & 29.1 & 31.6 \\
Qwen3-14B & 14B & 28.6 & 68.4 & 42.9 & 70.0 & 36.7 & 35.3 & 45.9 & 47.4 & 31.4 & 85.7 & 31.2 & 36.4 & 61.5 & 30.8 & 42.9 & 44.4 & 46.8 & 35.6 & 36.0 & 55.6 & 33.2 & 40.5 \\
gpt-oss-20b & 3.6B/21B & 75.0 & 47.4 & 42.9 & 70.0 & 54.2 & 35.3 & 43.2 & 47.4 & 43.5 & 42.9 & 25.4 & 42.4 & 46.2 & 38.5 & 43.3 & 39.3 & 49.6 & 40.7 & 30.9 & 55.6 & 50.0 & 43.3 \\
Qwen3-30B-A3B-Thinking-2507 & 3B/30B & 100.0 & 68.4 & 42.9 & 80.0 & 46.7 & 47.1 & 51.4 & 63.2 & 51.4 & 57.1 & 31.2 & 64.7 & 46.2 & 53.8 & 45.4 & 39.3 & 40.8 & 47.5 & 42.2 & 58.8 & 40.2 & 46.7 \\
Qwen3-30B-A3B-Instruct-2507 & 3B/30B & 50.0 & 68.4 & 42.9 & 60.0 & 36.7 & 35.3 & 51.4 & 63.2 & 34.3 & 57.1 & 21.9 & 41.2 & 38.5 & 46.2 & 42.9 & 53.6 & 45.1 & 39.5 & 37.8 & 44.4 & 40.0 & 42.3 \\
Qwen3-30B-A3B & 3B/30B & 50.0 & 42.1 & 57.1 & 60.0 & 43.3 & 35.3 & 49.3 & 57.9 & 42.9 & 57.1 & 25.0 & 29.4 & 46.2 & 53.8 & 46.3 & 50.0 & 40.8 & 33.3 & 32.5 & 35.3 & 43.4 & 41.0 \\
Qwen3-Coder-30B-A3B-Instruct & 3B/30B & 50.0 & 68.4 & 57.1 & 50.0 & 33.3 & 37.5 & 45.9 & 63.2 & 37.1 & 57.1 & 28.1 & 23.5 & 53.8 & 53.8 & 24.5 & 42.9 & 45.1 & 30.3 & 33.1 & 55.6 & 36.0 & 38.5 \\
\midrule\multicolumn{24}{c}{\textit{\textbf{32B+ Large Language Models}}} \\ \midrule
Qwen2.5-32B-Instruct & 32B & 75.0 & 52.6 & 42.9 & 60.0 & 36.7 & 29.4 & 37.8 & 47.4 & 17.1 & 14.3 & 25.0 & 35.3 & 32.0 & 46.2 & 39.2 & 39.3 & 45.1 & 35.0 & 29.4 & 35.3 & 39.4 & 36.7 \\
DeepSeek-R1-Distill-Qwen-32B & 32B & 75.0 & 42.1 & 14.3 & 30.0 & 30.0 & 17.6 & 29.7 & 59.5 & 31.9 & 28.6 & 21.9 & 25.0 & 23.1 & 48.0 & 41.2 & 28.6 & 34.0 & 30.1 & 29.6 & 47.1 & 35.4 & 32.8 \\
QwQ-32B & 32B & 75.0 & 63.2 & 30.8 & 40.0 & 40.0 & 41.2 & 41.7 & 75.7 & 50.0 & 71.4 & 36.1 & 41.2 & 64.0 & 61.5 & 45.8 & 32.7 & 51.1 & 35.8 & 33.8 & 47.1 & 43.7 & 44.0 \\
Qwen3-32B & 32B & 50.0 & 57.9 & 28.6 & 70.0 & 36.7 & 29.4 & 49.3 & 63.2 & 51.4 & 71.4 & 28.1 & 47.1 & 38.5 & 53.8 & 46.9 & 47.3 & 56.7 & 37.5 & 39.8 & 44.4 & 36.5 & 44.1 \\
Seed-OSS & 36B & 75.0 & 52.6 & 0.0 & 50.0 & 36.7 & 36.4 & 43.8 & 52.6 & 48.6 & 42.9 & 16.1 & 36.4 & 32.0 & 46.2 & 37.5 & 32.1 & 39.7 & 32.8 & 27.5 & 47.1 & 38.0 & 36.8 \\
DeepSeek-R1-Distill-Llama-70B & 70B & 75.0 & 48.6 & 15.4 & 30.0 & 36.7 & 29.4 & 32.4 & 47.4 & 49.3 & 28.6 & 25.4 & 41.2 & 46.2 & 38.5 & 41.7 & 28.6 & 46.5 & 34.7 & 28.0 & 35.3 & 44.0 & 38.0 \\
Qwen2.5-72B-Instruct & 72B & 50.0 & 63.2 & 30.8 & 50.0 & 30.0 & 29.4 & 43.2 & 63.2 & 42.9 & 42.9 & 21.9 & 42.4 & 23.1 & 53.8 & 41.2 & 50.0 & 45.7 & 40.7 & 30.7 & 44.4 & 41.4 & 40.5 \\
gpt-oss-120b & 5.1B/117B & 50.0 & 73.7 & 42.9 & 70.0 & 63.3 & 41.2 & 54.1 & 63.2 & 62.9 & 85.7 & 46.9 & 47.1 & 38.5 & 53.8 & 57.1 & 50.0 & 62.0 & 50.6 & 46.3 & 55.6 & 51.0 & 54.1 \\
Qwen3-235B-A22B-Instruct-2507 & 22B/235B & 75.0 & 68.4 & 42.9 & 70.0 & 53.3 & 35.3 & 56.8 & 78.9 & 51.4 & 57.1 & 28.1 & 50.0 & 46.2 & 53.8 & 49.0 & 60.7 & 56.3 & 48.3 & 43.9 & 55.6 & 42.0 & 49.9 \\
Qwen3-235B-A22B & 22B/235B & 50.0 & 52.6 & 42.9 & 80.0 & 43.3 & 35.3 & 51.4 & 68.4 & 48.6 & 71.4 & 31.2 & 36.4 & 38.5 & 38.5 & 49.5 & 39.3 & 57.7 & 42.0 & 31.9 & 55.6 & 39.0 & 44.5 \\
Qwen3-235B-A22B-Thinking-2507 & 22B/235B & 75.0 & 78.9 & 28.6 & 90.0 & 53.3 & 47.1 & 62.2 & 68.4 & 43.5 & 71.4 & 37.5 & 58.8 & 53.8 & 46.2 & 53.1 & 50.0 & 57.7 & 50.8 & 40.5 & 44.4 & 56.3 & 52.9 \\
Qwen3-Coder-480B-A35B-Instruct & 35B/480B & 75.0 & 68.4 & 42.9 & 70.0 & 50.0 & 36.4 & 48.6 & 64.9 & 37.7 & 57.1 & 34.4 & 35.3 & 61.5 & 61.5 & 49.5 & 47.3 & 60.6 & 45.2 & 42.0 & 55.6 & 40.0 & 47.7 \\
DeepSeek-V3 & 37B/671B & 75.0 & 57.9 & 42.9 & 80.0 & 46.7 & 52.9 & 51.4 & 73.7 & 48.6 & 71.4 & 40.6 & 54.5 & 46.2 & 61.5 & 44.9 & 46.4 & 57.7 & 46.1 & 40.2 & 44.4 & 48.2 & 49.6 \\
DeepSeek-V3-0324 & 37B/671B & 75.0 & 84.2 & 57.1 & 90.0 & 60.0 & 47.1 & 56.8 & 63.2 & 54.3 & 71.4 & 34.4 & 41.2 & 38.5 & 53.8 & 53.6 & 42.9 & 59.2 & 46.1 & 45.1 & 55.6 & 47.0 & 51.6 \\
DeepSeek-V3.1 & 37B/671B & 75.0 & 78.9 & 57.1 & 80.0 & 56.7 & 29.4 & 54.1 & 68.4 & 54.3 & 42.9 & 21.9 & 47.1 & 38.5 & 61.5 & 44.9 & 57.1 & 60.6 & 47.7 & 39.0 & 66.7 & 46.2 & 49.8 \\
DeepSeek-R1 & 37B/671B & 50.0 & 63.2 & 28.6 & 60.0 & 43.3 & 35.3 & 48.6 & 57.9 & 57.1 & 71.4 & 37.5 & 41.2 & 46.2 & 30.8 & 52.1 & 50.0 & 52.1 & 50.8 & 41.5 & 66.7 & 46.0 & 48.2 \\
DeepSeek-R1-0528 & 37B/671B & 50.0 & 47.4 & 28.6 & 70.0 & 50.0 & 35.3 & 54.1 & 68.4 & 51.4 & 57.1 & 25.0 & 52.9 & 46.2 & 69.2 & 53.1 & 39.3 & 49.3 & 47.2 & 39.3 & 44.4 & 48.2 & 47.5 \\
GLM-4.5 & 32B/355B & 50.0 & 63.2 & 42.9 & 90.0 & 53.3 & 41.2 & 51.4 & 63.2 & 54.3 & 71.4 & 25.0 & 41.2 & 53.8 & 46.2 & 51.0 & 46.4 & 56.7 & 52.0 & 47.9 & 58.8 & 49.2 & 50.9 \\
GLM-4.5-Air & 12B/106B & 50.0 & 63.2 & 42.9 & 90.0 & 46.7 & 5.9 & 48.6 & 52.6 & 45.7 & 57.1 & 28.1 & 47.1 & 38.5 & 38.5 & 49.0 & 29.1 & 56.3 & 46.1 & 34.1 & 58.8 & 42.4 & 44.3 \\ \midrule
Qwen2.5-Coder-32B-Instruct & 32B & 50.0 & 63.2 & 28.6 & 52.6 & 40.0 & 36.4 & 40.5 & 68.4 & 34.8 & 42.9 & 28.1 & 42.4 & 15.4 & 53.8 & 43.3 & 42.9 & 44.0 & 38.2 & 35.4 & 55.6 & 35.0 & 40.0 \\
\sftmodel{} & 32B & 50.0 & 70.6 & 50.0 & 41.7 & 46.7 & 41.2 & 38.5 & 52.4 & 42.9 & 28.6 & 27.3 & 41.2 & 20.0 & 53.8 & 31.4 & 64.3 & 48.6 & 39.3 & 38.4 & 55.6 & 42.2 & 42.3 \\
\rlmodel{} & 32B & 50.0 & 78.8 & 37.5 & 60.9 & 43.3 & 38.7 & 39.0 & 57.1 & 48.6 & 28.6 & 37.5 & 52.9 & 33.3 & 61.5 & 35.6 & 43.6 & 53.1 & 45.9 & 39.0 & 55.6 & 43.5 & 45.2 \\
\bottomrule
\end{tabular}}
\vspace{-5pt}
\caption{Chinese Results of the different domains in CodeSimpleQA (F1 Score). ``ADS'' denotes Algorithms and Data Structures, ``AI'' denotes Artificial Intelligence, ``BIO'' denotes Bioinformatics, ``CS'' denotes Computational Science, ``CG'' denotes Computer Graphics, ``CSYS'' denotes Computer Systems, ``CYB'' denotes Cybersecurity, ``DSA'' denotes Data Science and Analytics, ``DB'' denotes Databases, ``ET'' denotes Emerging Technologies, ``GD'' denotes Game Development, ``HCI'' denotes Human-Computer Interaction, ``IS'' denotes Information Systems, ``ML'' denotes Machine Learning, ``MC'' denotes Mobile Computing, ``NC'' denotes Networks and Communications, ``OS'' denotes Operating Systems, ``PL'' denotes Programming Languages, ``SE'' denotes Software Engineering, ``TOC'' denotes Theory of Computation, ``WT'' denotes Web Technologies, ``Avg.'' denotes Average.}
\vspace{-10pt}
\label{tab:chinese_results}
\end{table*}

\begin{table*}[t!]
\centering
\resizebox{1.0\textwidth}{!}{
\begin{tabular}{l|c|ccccccccccccccccccccc|c}
\toprule
Model & Params & ADS & AI & BIO & CS & CG & CSYS & CYB & DSA & DB & ET & GD & HCI & IS & ML & MC & NC & OS & PL & SE & TOC & WT & Avg. \\
\midrule
\multicolumn{24}{c}{\textit{\textbf{32B+ Large Language Models}}} \\ \midrule
gpt-4o-2024-11-20 & \faLock{} & 28.6 & 50.0 & 33.3 & 73.7 & 50.0 & 33.3 & 62.5 & 42.3 & 39.5 & 66.7 & 68.4 & 54.5 & 20.0 & 0.0 & 68.0 & 35.3 & 44.4 & 51.5 & 36.9 & 50.0 & 45.5 & 46.7 \\
chatgpt-4o-latest & \faLock{} & 60.0 & 50.0 & 0.0 & 63.6 & 58.8 & 30.0 & 62.5 & 40.6 & 48.3 & 75.0 & 66.7 & 59.3 & 23.1 & 25.0 & 63.9 & 45.0 & 43.1 & 58.1 & 36.4 & 0.0 & 51.4 & 50.5 \\
o3 & \faLock{} & 40.0 & 33.3 & 50.0 & 72.7 & 82.4 & 60.0 & 70.8 & 59.4 & 69.0 & 66.7 & 57.7 & 78.6 & 30.8 & 62.5 & 72.2 & 65.0 & 60.8 & 62.5 & 54.3 & 100.0 & 58.0 & 61.2 \\
o4-mini & \faLock{} & 40.0 & 50.0 & 25.0 & 60.0 & 58.8 & 30.0 & 50.0 & 56.7 & 57.9 & 50.0 & 68.0 & 42.9 & 38.5 & 50.0 & 65.7 & 40.0 & 51.1 & 55.8 & 39.3 & 100.0 & 47.9 & 51.1 \\
claude-sonnet-4-20250514-thinking & \faLock{} & 44.4 & 25.0 & 50.0 & 50.0 & 36.4 & 61.5 & 58.3 & 54.5 & 40.4 & 35.3 & 44.4 & 50.0 & 36.4 & 46.2 & 45.3 & 62.5 & 49.3 & 55.8 & 43.0 & 100.0 & 53.2 & 49.6 \\
gpt-5-mini & \faLock{} & 50.0 & 60.0 & 50.0 & 70.6 & 53.8 & 57.1 & 47.1 & 50.0 & 40.5 & 50.0 & 54.5 & 75.0 & 41.7 & 25.0 & 68.2 & 50.0 & 53.7 & 58.4 & 46.7 & 100.0 & 56.5 & 53.5 \\
o3-mini-2025-01-31 & \faLock{} & 50.0 & 66.7 & 50.0 & 66.7 & 52.9 & 40.0 & 54.2 & 50.0 & 48.3 & 58.3 & 57.7 & 50.0 & 30.8 & 62.5 & 58.3 & 40.0 & 35.3 & 51.4 & 37.2 & 100.0 & 45.2 & 47.5 \\
gpt-5 & \faLock{} & 35.3 & 20.0 & 100.0 & 66.7 & 69.2 & 80.0 & 62.9 & 78.8 & 48.8 & 81.8 & 57.1 & 100.0 & 40.0 & 71.4 & 65.2 & 66.7 & 54.3 & 68.4 & 58.0 & 100.0 & 64.2 & 62.9 \\
claude-sonnet-4-20250514 & \faLock{} & 33.3 & 66.7 & 50.0 & 35.3 & 44.4 & 36.4 & 61.5 & 57.1 & 50.0 & 55.6 & 50.0 & 63.6 & 16.7 & 61.5 & 68.1 & 62.5 & 51.9 & 53.3 & 37.1 & 0.0 & 51.6 & 50.3 \\
\midrule\multicolumn{24}{c}{\textit{\textbf{32B+ Large Language Models}}} \\ \midrule
Qwen2.5-0.5B-Instruct & 0.5B & 21.1 & 0.0 & 0.0 & 36.4 & 12.1 & 0.0 & 8.7 & 3.2 & 1.8 & 8.7 & 3.9 & 7.4 & 0.0 & 26.7 & 22.9 & 5.1 & 12.0 & 15.5 & 1.1 & 50.0 & 14.9 & 11.0 \\
Qwen2.5-Coder-0.5B-Instruct & 0.5B & 21.1 & 20.0 & 0.0 & 18.2 & 17.6 & 0.0 & 9.1 & 3.2 & 3.6 & 9.5 & 7.8 & 7.1 & 8.0 & 0.0 & 39.4 & 0.0 & 22.4 & 16.8 & 5.7 & 0.0 & 16.5 & 13.5 \\
Qwen3-0.6B & 0.6B & 20.0 & 16.7 & 0.0 & 0.0 & 17.6 & 0.0 & 12.8 & 6.3 & 10.7 & 8.3 & 8.2 & 7.1 & 0.0 & 0.0 & 31.0 & 15.0 & 18.0 & 11.7 & 3.3 & 50.0 & 13.8 & 11.7 \\
Llama-3.2-1B-Instruct & 1B & 0.0 & 33.3 & 0.0 & 0.0 & 23.5 & 0.0 & 18.2 & 9.7 & 7.0 & 19.0 & 4.1 & 0.0 & 8.0 & 25.0 & 22.9 & 0.0 & 13.9 & 10.7 & 4.5 & 0.0 & 14.1 & 10.9 \\
Qwen2.5-Coder-1.5B-Instruct & 1.5B & 20.0 & 16.7 & 0.0 & 28.6 & 12.1 & 0.0 & 30.4 & 19.0 & 8.8 & 16.7 & 15.4 & 21.4 & 0.0 & 0.0 & 44.4 & 10.0 & 25.7 & 22.1 & 10.8 & 100.0 & 24.4 & 20.1 \\
Qwen2.5-1.5B-Instruct & 1.5B & 20.0 & 16.7 & 0.0 & 18.2 & 23.5 & 0.0 & 20.8 & 9.4 & 12.1 & 25.0 & 15.4 & 14.8 & 0.0 & 25.0 & 33.3 & 5.3 & 15.8 & 21.4 & 4.3 & 50.0 & 25.2 & 17.8 \\
DeepSeek-R1-Distill-Qwen-1.5B & 1.5B & 20.0 & 16.7 & 0.0 & 0.0 & 24.2 & 0.0 & 12.8 & 9.5 & 3.6 & 8.7 & 3.8 & 0.0 & 0.0 & 0.0 & 11.3 & 0.0 & 2.0 & 14.6 & 2.1 & 0.0 & 5.9 & 7.4 \\
Llama-3.2-3B-Instruct & 3B & 0.0 & 16.7 & 0.0 & 20.0 & 23.5 & 0.0 & 22.2 & 3.4 & 14.3 & 40.0 & 8.9 & 14.3 & 0.0 & 40.0 & 34.3 & 0.0 & 16.0 & 19.0 & 8.2 & 0.0 & 16.5 & 15.6 \\
Qwen2.5-Coder-3B-Instruct & 3B & 20.0 & 54.5 & 0.0 & 27.3 & 30.3 & 10.0 & 37.5 & 18.8 & 15.7 & 26.1 & 11.8 & 21.4 & 0.0 & 25.0 & 44.4 & 15.0 & 23.8 & 25.7 & 15.0 & 50.0 & 16.9 & 21.3 \\
Qwen2.5-3B-Instruct & 3B & 20.0 & 18.2 & 0.0 & 0.0 & 17.6 & 20.0 & 29.8 & 15.6 & 13.8 & 16.7 & 19.6 & 14.3 & 0.0 & 0.0 & 33.3 & 10.3 & 20.0 & 22.2 & 9.7 & 100.0 & 21.5 & 18.6 \\
Qwen3-4B & 4B & 21.1 & 16.7 & 0.0 & 38.1 & 17.6 & 20.0 & 25.5 & 22.2 & 19.1 & 50.0 & 28.0 & 35.7 & 0.0 & 12.5 & 42.9 & 10.5 & 35.6 & 30.1 & 13.0 & 50.0 & 28.7 & 26.0 \\
Qwen3-4B-Thinking-2507 & 4B & 30.0 & 33.3 & 25.0 & 27.3 & 23.5 & 20.0 & 25.0 & 22.2 & 24.6 & 16.7 & 38.5 & 28.6 & 8.0 & 0.0 & 31.0 & 20.0 & 21.6 & 31.1 & 12.9 & 50.0 & 25.7 & 24.7 \\
DeepSeek-R1-Distill-Qwen-7B & 7B & 20.0 & 0.0 & 0.0 & 18.2 & 18.2 & 0.0 & 20.8 & 12.5 & 8.7 & 0.0 & 19.2 & 0.0 & 0.0 & 0.0 & 13.9 & 0.0 & 14.0 & 17.6 & 4.3 & 50.0 & 13.3 & 12.1 \\
\midrule\multicolumn{24}{c}{\textit{\textbf{32B+ Large Language Models}}} \\ \midrule
Qwen2.5-Coder-7B-Instruct & 7B & 40.0 & 16.7 & 0.0 & 9.1 & 17.6 & 20.0 & 42.6 & 21.9 & 24.1 & 16.7 & 34.6 & 29.6 & 8.0 & 12.5 & 44.4 & 25.0 & 26.0 & 29.5 & 15.0 & 100.0 & 23.2 & 25.2 \\
Qwen2.5-7B-Instruct & 7B & 20.0 & 16.7 & 0.0 & 19.0 & 35.3 & 20.0 & 34.0 & 28.1 & 27.6 & 16.7 & 26.9 & 29.6 & 7.7 & 0.0 & 36.1 & 10.0 & 23.8 & 22.7 & 16.0 & 100.0 & 22.7 & 23.1 \\
Qwen3-1.7B & 7B & 21.1 & 50.0 & 0.0 & 18.2 & 23.5 & 20.0 & 29.2 & 21.9 & 25.9 & 25.0 & 27.5 & 21.4 & 0.0 & 12.5 & 33.3 & 0.0 & 19.8 & 23.3 & 11.7 & 100.0 & 19.8 & 20.8 \\
Hunyuan-7B-Instruct & 7B & 20.0 & 0.0 & 0.0 & 9.5 & 17.6 & 20.0 & 34.0 & 9.4 & 10.6 & 8.7 & 12.2 & 7.1 & 15.4 & 0.0 & 29.0 & 15.4 & 11.9 & 20.5 & 7.5 & 0.0 & 18.7 & 15.9 \\
Llama-3.1-8B-Instruct & 8B & 31.6 & 33.3 & 25.0 & 38.1 & 23.5 & 10.0 & 36.4 & 16.1 & 23.2 & 27.3 & 8.9 & 21.4 & 7.7 & 40.0 & 50.7 & 5.1 & 32.0 & 29.6 & 14.6 & 100.0 & 21.0 & 24.4 \\
DeepSeek-R1-Distill-Llama-8B & 8B & 20.0 & 0.0 & 0.0 & 9.1 & 11.8 & 10.0 & 25.0 & 18.8 & 15.5 & 25.0 & 19.2 & 14.3 & 0.0 & 0.0 & 27.8 & 10.0 & 17.6 & 19.4 & 6.4 & 50.0 & 15.1 & 15.6 \\
Seed-Coder-8B-Instruct & 8B & 40.0 & 18.2 & 0.0 & 18.2 & 11.8 & 10.0 & 37.5 & 12.5 & 25.9 & 25.0 & 26.9 & 29.6 & 0.0 & 0.0 & 33.3 & 30.0 & 23.8 & 26.5 & 10.8 & 0.0 & 29.6 & 23.8 \\
Qwen3-8B & 8B & 40.0 & 36.4 & 0.0 & 18.2 & 23.5 & 10.0 & 33.3 & 29.0 & 22.6 & 34.8 & 28.6 & 28.6 & 0.0 & 0.0 & 37.1 & 45.0 & 29.7 & 32.7 & 9.8 & 50.0 & 30.3 & 27.1 \\
DeepSeek-R1-0528-Qwen3-8B & 8B & 11.1 & 18.2 & 0.0 & 36.4 & 24.2 & 10.0 & 39.1 & 13.8 & 18.3 & 26.1 & 31.4 & 30.8 & 0.0 & 0.0 & 39.4 & 27.0 & 25.0 & 30.3 & 12.6 & 66.7 & 27.9 & 24.8 \\
Qwen2.5-Coder-14B-Instruct & 14B & 40.0 & 16.7 & 0.0 & 36.4 & 29.4 & 10.0 & 45.8 & 21.9 & 24.1 & 50.0 & 19.2 & 51.9 & 15.4 & 12.5 & 50.0 & 30.0 & 30.0 & 32.8 & 19.3 & 50.0 & 32.1 & 30.0 \\
Qwen2.5-14B-Instruct & 14B & 40.0 & 0.0 & 0.0 & 27.3 & 23.5 & 30.0 & 33.3 & 28.1 & 27.6 & 33.3 & 42.3 & 29.6 & 7.7 & 25.0 & 38.9 & 20.5 & 25.7 & 27.8 & 12.9 & 100.0 & 30.6 & 27.2 \\
DeepSeek-R1-Distill-Qwen-14B & 14B & 20.0 & 16.7 & 0.0 & 9.1 & 17.6 & 0.0 & 37.5 & 25.0 & 24.1 & 16.7 & 31.4 & 7.1 & 8.0 & 12.5 & 38.9 & 20.0 & 23.5 & 25.8 & 13.8 & 50.0 & 24.9 & 23.0 \\
Qwen3-14B & 14B & 40.0 & 36.4 & 25.0 & 36.4 & 23.5 & 10.0 & 41.7 & 31.7 & 26.3 & 41.7 & 52.0 & 35.7 & 7.7 & 0.0 & 50.7 & 50.0 & 23.5 & 31.1 & 21.6 & 50.0 & 31.1 & 30.9 \\
gpt-oss-20b & 3.6B/21B & 50.0 & 16.7 & 28.6 & 27.3 & 17.6 & 20.0 & 38.3 & 28.1 & 36.2 & 43.5 & 57.7 & 28.6 & 23.1 & 37.5 & 47.2 & 30.0 & 39.6 & 36.7 & 27.8 & 50.0 & 38.0 & 35.8 \\
GLM-4.5-Air & 12B/102B & 44.4 & 36.4 & 0.0 & 57.1 & 24.2 & 40.0 & 31.1 & 39.3 & 33.0 & 26.1 & 34.0 & 51.9 & 24.0 & 13.3 & 58.0 & 30.8 & 38.3 & 37.5 & 26.2 & 100.0 & 39.5 & 36.5 \\
Qwen3-30B-A3B-Thinking-2507 & 30B & 30.0 & 16.7 & 25.0 & 36.4 & 23.5 & 20.0 & 34.0 & 28.6 & 27.6 & 41.7 & 43.1 & 50.0 & 16.0 & 0.0 & 44.4 & 35.0 & 31.7 & 32.2 & 16.0 & 50.0 & 35.4 & 30.9 \\
Qwen3-30B-A3B-Instruct-2507 & 30B & 40.0 & 33.3 & 0.0 & 36.4 & 29.4 & 30.0 & 45.8 & 25.0 & 34.5 & 58.3 & 42.3 & 50.0 & 38.5 & 12.5 & 52.8 & 35.0 & 31.4 & 36.9 & 17.0 & 100.0 & 37.1 & 34.7 \\
Qwen3-30B-A3B & 30B & 40.0 & 33.3 & 0.0 & 45.5 & 23.5 & 20.0 & 37.5 & 22.6 & 31.3 & 41.7 & 43.1 & 64.3 & 23.1 & 0.0 & 56.3 & 35.0 & 37.3 & 33.7 & 16.3 & 100.0 & 33.5 & 32.7 \\
Qwen3-Coder-30B-A3B-Instruct & 30B & 42.1 & 50.0 & 0.0 & 38.1 & 35.3 & 20.0 & 38.3 & 34.9 & 29.6 & 8.7 & 29.2 & 57.1 & 16.0 & 0.0 & 39.4 & 30.0 & 33.3 & 33.8 & 23.9 & 100.0 & 30.4 & 31.3 \\
\midrule\multicolumn{24}{c}{\textit{\textbf{32B+ Large Language Models}}} \\ \midrule
Qwen2.5-32B-Instruct & 32B & 30.0 & 16.7 & 25.0 & 47.6 & 41.2 & 30.0 & 29.2 & 28.6 & 34.5 & 26.1 & 42.3 & 28.6 & 0.0 & 25.0 & 50.0 & 15.0 & 30.0 & 39.2 & 17.2 & 100.0 & 33.6 & 32.3 \\
DeepSeek-R1-Distill-Qwen-32B & 32B & 30.0 & 33.3 & 0.0 & 36.4 & 23.5 & 20.0 & 33.3 & 12.5 & 24.1 & 26.1 & 34.6 & 21.4 & 0.0 & 0.0 & 52.8 & 15.4 & 31.4 & 29.5 & 19.1 & 50.0 & 30.1 & 27.2 \\
QwQ-32B & 32B & 40.0 & 36.4 & 25.0 & 63.6 & 29.4 & 30.0 & 41.7 & 34.9 & 28.6 & 17.4 & 34.6 & 42.9 & 15.4 & 0.0 & 53.5 & 30.0 & 29.7 & 36.2 & 22.6 & 50.0 & 39.8 & 34.4 \\
Qwen3-32B & 32B & 30.0 & 33.3 & 25.0 & 57.1 & 29.4 & 20.0 & 42.6 & 50.0 & 24.3 & 41.7 & 47.1 & 35.7 & 15.4 & 37.5 & 37.1 & 25.6 & 31.7 & 45.9 & 19.4 & 50.0 & 37.0 & 35.6 \\
Seed-OSS & 36B & 30.0 & 16.7 & 25.0 & 27.3 & 29.4 & 20.0 & 33.3 & 34.9 & 26.1 & 16.7 & 29.2 & 28.6 & 24.0 & 0.0 & 42.3 & 30.0 & 35.3 & 33.1 & 25.7 & 50.0 & 32.7 & 30.7 \\
DeepSeek-R1-Distill-Llama-70B & 70B & 20.0 & 33.3 & 25.0 & 36.4 & 17.6 & 40.0 & 41.7 & 31.2 & 37.9 & 25.0 & 35.3 & 35.7 & 7.7 & 12.5 & 50.0 & 30.0 & 35.3 & 34.6 & 23.5 & 50.0 & 29.6 & 31.9 \\
Qwen2.5-72B-Instruct & 72B & 40.0 & 33.3 & 25.0 & 45.5 & 41.2 & 10.0 & 45.8 & 34.4 & 37.9 & 41.7 & 46.2 & 37.0 & 7.7 & 50.0 & 52.8 & 25.0 & 37.6 & 44.9 & 28.9 & 50.0 & 40.1 & 38.9 \\
gpt-oss-120b & 5.1B/117B & 40.0 & 50.0 & 0.0 & 54.5 & 29.4 & 40.0 & 45.8 & 53.1 & 50.0 & 41.7 & 61.5 & 64.3 & 23.1 & 37.5 & 52.8 & 35.0 & 52.9 & 51.2 & 34.0 & 100.0 & 46.2 & 46.6 \\ 
Qwen3-235B-A22B-Instruct-2507 & 235B & 42.1 & 50.0 & 0.0 & 54.5 & 47.1 & 40.0 & 41.7 & 37.5 & 44.8 & 58.3 & 42.3 & 35.7 & 30.8 & 37.5 & 52.8 & 40.0 & 35.3 & 43.3 & 34.0 & 100.0 & 40.6 & 41.2 \\
Qwen3-235B-A22B & 235B & 50.0 & 50.0 & 25.0 & 63.6 & 29.4 & 40.0 & 29.2 & 50.8 & 36.2 & 58.3 & 50.0 & 42.9 & 23.1 & 25.0 & 45.1 & 55.0 & 39.2 & 49.1 & 31.0 & 100.0 & 40.8 & 42.0 \\
Qwen3-235B-A22B-Thinking-2507 & 235B & 50.0 & 50.0 & 0.0 & 54.5 & 58.8 & 40.0 & 50.0 & 34.4 & 43.1 & 41.7 & 61.5 & 50.0 & 30.8 & 25.0 & 52.8 & 30.0 & 25.5 & 50.2 & 32.1 & 50.0 & 41.7 & 42.5 \\
Qwen3-Coder-480B-A35B-Instruct & 480B & 50.0 & 33.3 & 0.0 & 28.6 & 52.9 & 30.0 & 46.8 & 19.4 & 40.0 & 58.3 & 39.1 & 64.3 & 48.0 & 40.0 & 70.4 & 45.0 & 47.1 & 44.8 & 28.0 & 100.0 & 43.7 & 42.6 \\
Qwen3-Coder-480B & 480B & 30.0 & 18.2 & 0.0 & 11.1 & 43.7 & 27.3 & 42.9 & 13.8 & 35.5 & 60.9 & 34.8 & 64.3 & 34.8 & 42.9 & 64.5 & 32.4 & 33.1 & 39.9 & 32.6 & 100.0 & 42.4 & 38.2 \\
DeepSeek-R1 & 37B/671B & 40.0 & 33.3 & 50.0 & 54.5 & 35.3 & 30.0 & 54.2 & 50.0 & 41.4 & 41.7 & 61.5 & 42.9 & 38.5 & 12.5 & 47.2 & 55.0 & 51.0 & 53.5 & 31.9 & 100.0 & 46.2 & 46.2 \\
DeepSeek-R1-0528 & 37B/671B & 40.0 & 50.0 & 25.0 & 54.5 & 35.3 & 40.0 & 58.3 & 50.0 & 36.2 & 50.0 & 53.8 & 57.1 & 46.2 & 37.5 & 61.1 & 40.0 & 37.3 & 50.6 & 32.4 & 100.0 & 42.1 & 44.5 \\
DeepSeek-V3 & 37B/671B & 40.0 & 33.3 & 25.0 & 54.5 & 29.4 & 20.0 & 58.3 & 46.9 & 43.1 & 66.7 & 46.2 & 50.0 & 30.8 & 12.5 & 52.8 & 40.0 & 33.3 & 45.6 & 26.6 & 50.0 & 40.0 & 40.7 \\
DeepSeek-V3-0324 & 37B/671B & 40.0 & 33.3 & 50.0 & 63.6 & 41.2 & 40.0 & 54.2 & 50.0 & 36.2 & 66.7 & 50.0 & 64.3 & 23.1 & 12.5 & 50.0 & 50.0 & 37.3 & 44.4 & 35.1 & 50.0 & 44.5 & 43.4 \\
DeepSeek-V3.1 & 37B/671B & 30.0 & 0.0 & 50.0 & 54.5 & 41.2 & 30.0 & 45.8 & 46.9 & 43.1 & 50.0 & 38.5 & 50.0 & 15.4 & 25.0 & 47.2 & 20.0 & 41.6 & 49.4 & 30.9 & 50.0 & 38.2 & 40.5 \\
GLM-4.5 & 32B/355B & 40.0 & 66.7 & 25.0 & 66.7 & 29.4 & 40.0 & 45.8 & 66.7 & 43.1 & 50.0 & 57.7 & 50.0 & 30.8 & 12.5 & 47.2 & 50.0 & 35.6 & 47.5 & 41.9 & 50.0 & 43.1 & 45.0 \\\midrule
Qwen2.5-Coder-32B-Instruct & 32B & 40.0 & 16.7 & 0.0 & 54.5 & 47.1 & 10.0 & 37.5 & 28.1 & 31.3 & 33.3 & 39.2 & 51.9 & 7.7 & 25.0 & 52.8 & 25.0 & 29.4 & 37.5 & 26.6 & 100.0 & 33.5 & 33.9 \\
\sftmodel{} & 32B & 36.4 & 25.0 & 0.0 & 63.6 & 41.2 & 20.0 & 33.3 & 34.4 & 34.5 & 16.7 & 53.8 & 51.9 & 15.4 & 25.0 & 66.7 & 25.0 & 33.7 & 41.6 & 34.2 & 50.0 & 41.7 & 39.2 \\
\rlmodel{} & 32B & 28.6 & 50.0 & 0.0 & 63.6 & 35.3 & 20.0 & 37.5 & 31.7 & 43.1 & 33.3 & 58.8 & 50.0 & 40.0 & 26.7 & 62.0 & 45.0 & 37.6 & 46.5 & 29.9 & 100.0 & 44.6 & 42.3 \\
\bottomrule
\end{tabular}}
\vspace{-5pt}
\caption{English Results of the different domains in CodeSimpleQA (F1 Score).}
\label{tab:english_results}
\vspace{-15pt}
\end{table*}

\section{Experiments}
\paragraph{LLMs}
The LLMs to be tested span multiple families including OpenAI's o-series (o3, o4-mini, o3-mini) and GPT variants (gpt-5, gpt-4o), the Claude Sonnet 4 models of the Anthropic, the reasoning model Deepseek-R1 series and their V3/V3.1 architectures, Alibaba's extensive Qwen2.5 and Qwen3 from Alibaba lineups (ranging from 0.5B to 480B parameters in base, instruct, coder, and thinking configurations), Meta's Llama 3.1 and 3.2 models, Zhipu AI's GLM-4.5 variants, and specialized models like QwQ-32B and various distilled versions optimized for efficiency. 

\paragraph{Implementation Details}
We fine-tune Qwen2.5-Coder-32B-Instruct on 66 Million \instruct{} samples using 32 NVIDIA H20 GPUs\footnote{\url{https://github.com/NVIDIA/Megatron-LM}} with Adam optimizer and cosine-decay scheduler (100 warmup steps, peak learning rate 6e-5, global batch size 1,024, tensor parallel size 4, 8,192 token limit). For RL, we train the RFT model \sftmodel{} using the VeRL~\cite{hybridflow} on 64 GPUs in FSDP mode with vLLM~\citep{vllm} as the inference backend, using constant learning rate 5e-7, batch size 1,024 queries, and 8 trajectories per group.

\paragraph{Evaluation Metrics.}
5 metrics are reported on \benchmark{}~\cite{c_simpleqa}, including (1) Correct (CO): The predicted answer fully includes the reference answer without introducing any contradictory elements. (2) Not attempted (NA): The reference answer is not fully given in the predicted answer, and there are no contradictory elements with the reference answer. (3) Incorrect (IN): The predicted answer contradicts the reference answer, even if the contradiction is solved. (4) Correct given attempted (CGA): The metric is the proportion of accurately answered questions among the attempted questions.
(5) F-score: the harmonic mean between correct and correct given attempted. 

\begin{figure*}[htbp]
\begin{center}
    \includegraphics[width=0.9\textwidth]{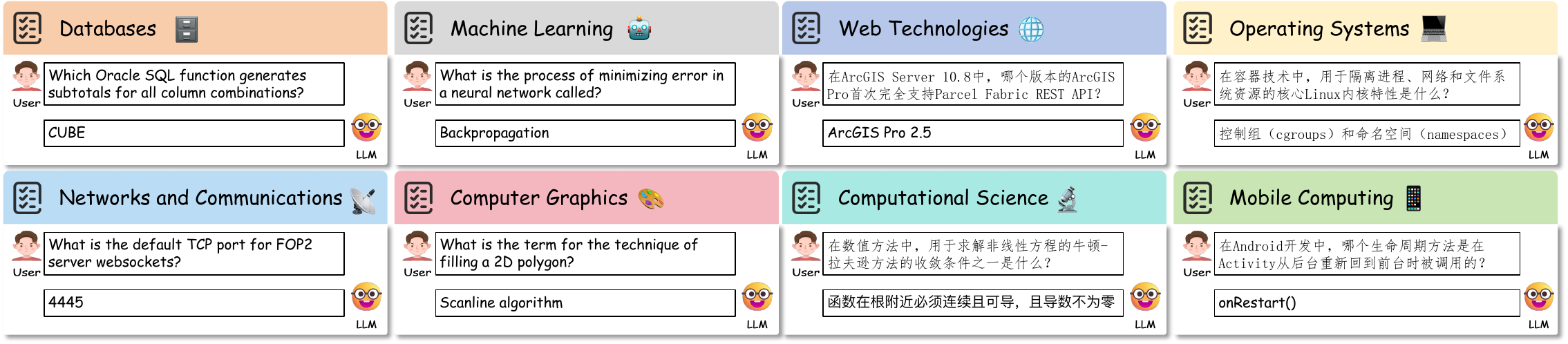}
    \vspace{-5pt}
    \caption{Domains statistics in \benchmark{}.}
    \label{fig:examples}
    \vspace{-15pt}
\end{center}
\end{figure*}
\begin{figure*}[htbp]
\begin{center}
    \includegraphics[width=0.85\textwidth]{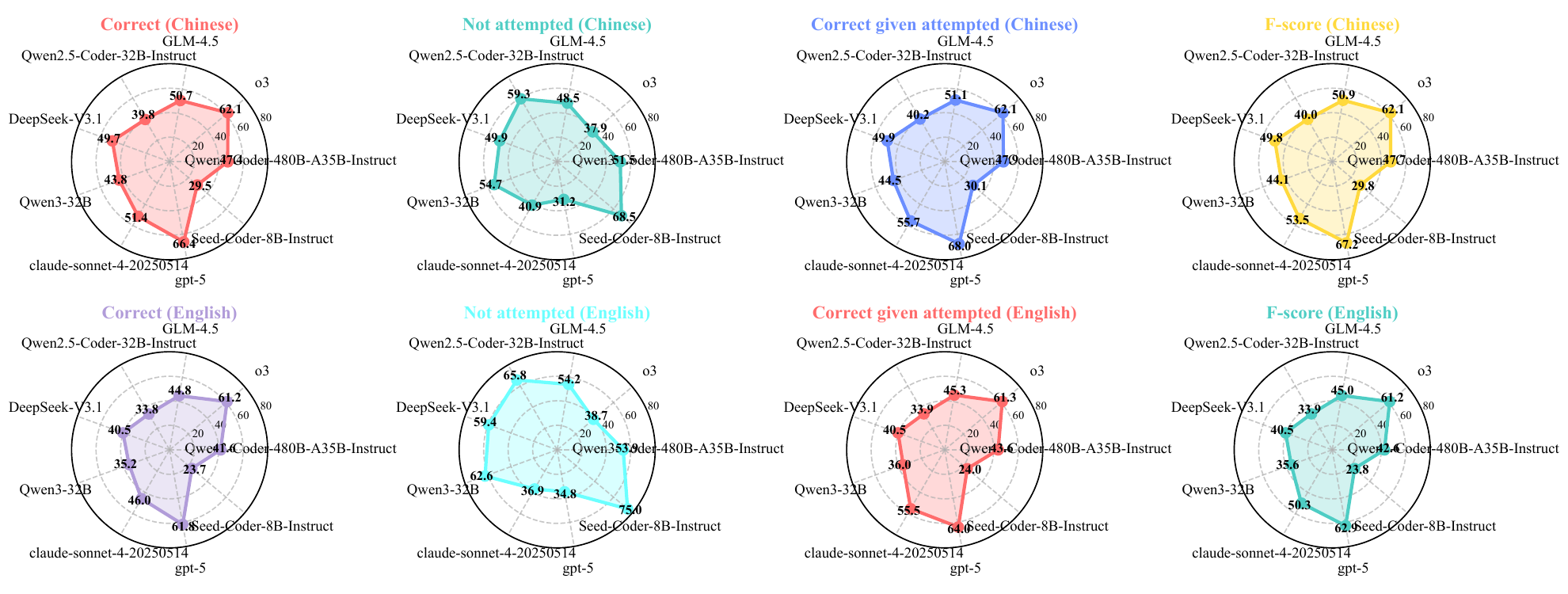}
    \caption{Results of different metrics in \benchmark{}.}
    \label{fig:diff_metrics}
    \vspace{-25pt}
\end{center}
\end{figure*}

\paragraph{Chinese Results on Different Domains.}
In \autoref{tab:chinese_results}, the Chinese evaluation reveal significant performance variations across models and domains. Proprietary LLMs dominate the leaderboard, with gpt-5 achieving the highest average F-score, followed by o3 and gpt-5-mini. Among thinking LLMs, claude-sonnet-4-20250514-thinking reaches 55.4\%, demonstrating the benefit of extended reasoning for factual code questions. Open-source LLMs show more varied performance, with the largest models like GLM-4.5 (50.9\%) approaching but not matching proprietary model capabilities. Our CodeSimpleQA-RL model achieves 45.2\%, outperforming the baseline Qwen2.5-Coder-32B-Instruct by a large margin, validating the effectiveness of RL for factual code QA. Smaller LLMs (0.5B$\sim$3B) consistently struggle, indicating that factual code knowledge requires sufficient model capacity. Domain-specific analysis reveals that models generally perform better on web technologies, software engineering, and programming languages, while struggling with specialized areas like theory of computation and computer graphics.

\paragraph{English Results on Different Domains.}
\autoref{tab:english_results} shows similar trends but differs from Chinese performance. GPT-5 maintains its lead with 62.9\%, closely followed by o3. Claude-sonnet-4-20250514-thinking achieves 49.6\%, though slightly lower than its Chinese counterpart. Open-source LLMs like DeepSeek-R1 show competitive results, again surpassing Qwen2.5-Coder-32B-Instruct by 8.4 percentage points, a larger improvement than observed in Chinese. English results exhibit slightly lower scores across most models compared to Chinese, likely due to training data differences. Domain-specific patterns remain consistent, with strong performance on Databases, Web Technologies, and Software Engineering, while specialized domains like Bioinformatics and Theory of Computation continue to challenge top models. Cross-lingual comparison suggests that code knowledge evaluation benefits from bilingual assessment to capture model capabilities.

\section{Analysis}
\paragraph{Example Study.}
\autoref{fig:examples} lists multiple examples from \benchmark{} spanning diverse computer science domains, showcasing eight technical categories with concise factual answers. The dataset demonstrates breadth across topics ranging from \textcolor{blue}{Oracle SQL functions} to \textcolor{blue}{graphics algorithms}. \benchmark{} contains both English and Chinese questions, such as queries about ArcGIS Server APIs, numerical differentiation methods, and Android lifecycle methods (\textcolor{orange}{onRestart()}), providing a benchmark for assessing understanding of fundamental computer science concepts.
\begin{figure*}[htbp]
\begin{center}
    \includegraphics[width=0.75\textwidth]{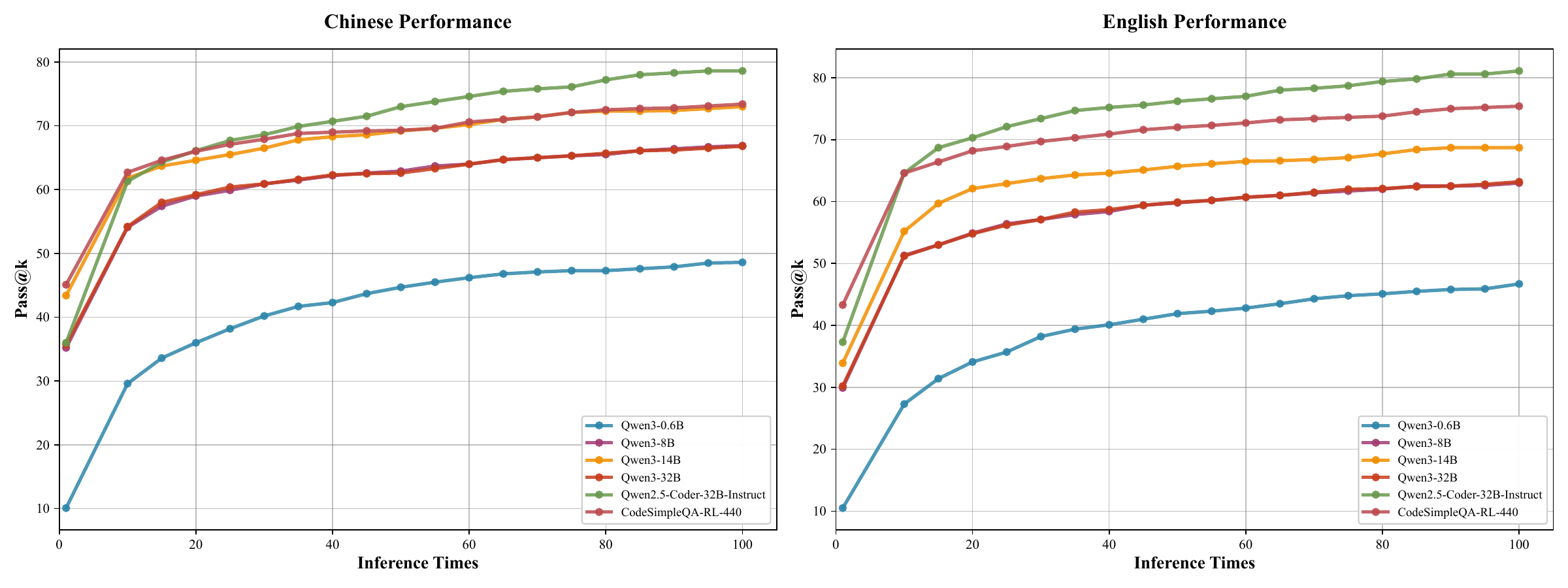}
    \vspace{-5pt}
    \caption{Results of Pass@k in \benchmark{} with increasing samples.}
    \label{fig:pass_at_k}
    \vspace{-15pt}
\end{center}
\end{figure*}
\begin{figure*}[thbp]
\begin{center}
    \includegraphics[width=0.75\textwidth]{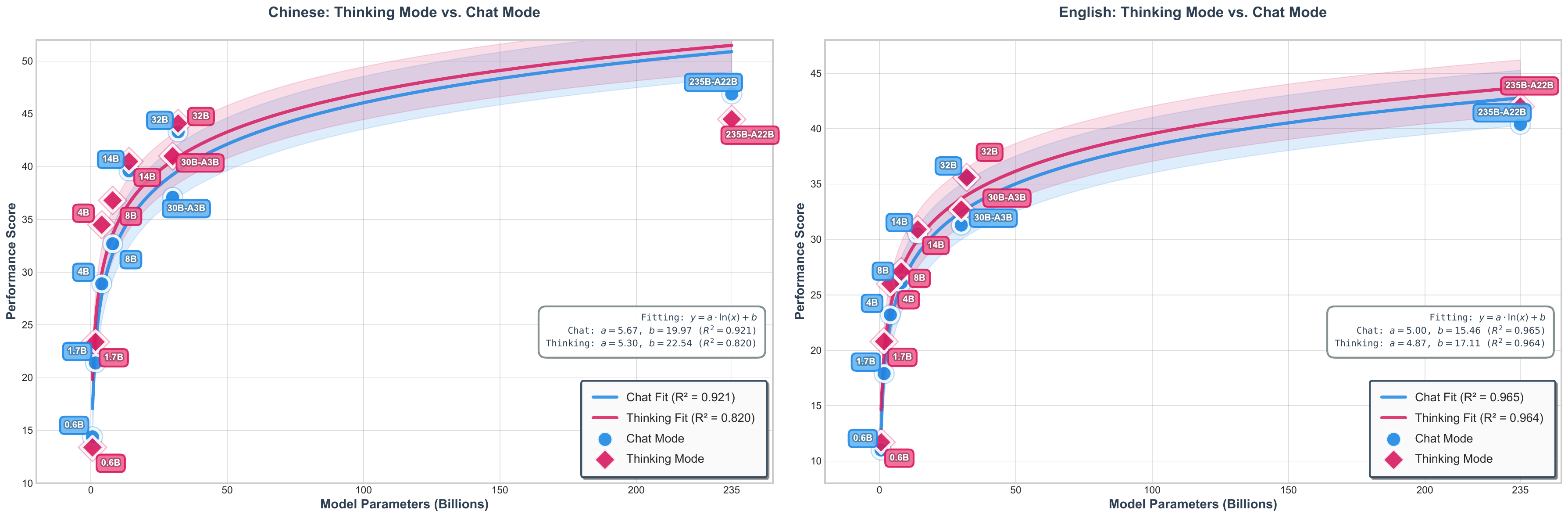}
    \vspace{-2pt}
    \caption{Comparison between non-thinking and thinking mode.}
    \label{fig:non_thinking_and_thinking}
    \vspace{-20pt}
\end{center}
\end{figure*}

\paragraph{Results of different metrics.}
\autoref{fig:diff_metrics} presents a comprehensive evaluation comparing six large language models (GLM-4.5, GLM-4-Plus, Qwen2.5-Coder-32B-Instruct, DeepSeek-V3, Qwen3 variants, and Claude-Sonnet-4) across both Chinese and English languages using four distinct metrics displayed as radar plots. Deepseek-V3.1 gets the best performance, while the small Seed-Coder (8B) tends not to answer questions (Not Attempted).

\paragraph{Comparison between RAG method and SFT.}
We compare RAG with large-scale SFT on \benchmark{}, where RAG augments answers by retrieving relevant documentation at inference time, while SFT internalizes code-related factual knowledge by optimizing parameters with QA pairs. 
We create a subset test after 2024 (100 samples), and a subset test before 2024 (100 samples), and the results show RAG achieves 71.0/68.0 (old/new) scores compared to \rlmodel{} (42.0/24.0), but it depends heavily on external knowledge base quality. SFT requires high-quality instruction data, yet provides faster inference and more consistent responses. These results indicate that combining SFT for core knowledge with RAG for up-to-date technical information may work best in production environments.

\paragraph{Test Time Scaling.}
\autoref{fig:pass_at_k} lists the varying Pass@k as inference times increase from 1 to 100. All LLMs exhibit rapid performance gains during the initial 10-20 inference iterations before gradually plateauing, where the larger LLMs demonstrate superior performance and the performance hierarchy among LLMs remains relatively stable.

\paragraph{Comparison between Chat and Thinking mode.}
\autoref{fig:non_thinking_and_thinking} presents scatter plots analyzing the relationship between model parameters and performance scores for LLMs in Chinese (left) and English (right), specifically comparing chat mode (blue line) and thinking (red line) mode. The thinking mode consistently outperforms the chat mode, with the performance gap maintained across the entire range of model sizes from small LLMs (under 10B parameters) to large LLMs (over 200B parameters). The regression fits show strong correlations ($R^{2}$ values around $0.92\sim 0.96$), indicating that performance scales predictably with model size $y = a \cdot \text{ln}(x) + b$. The parallel upward trends in both modes suggest that larger LLMs perform better overall, enabling the reasoning capabilities to provide a consistent performance advantage.

\section{Related Work}
\paragraph{Code Post-training.} 
Code post-training~\cite{deepseek_coder,qwen25coder,magicoder,gpt4,gpt5,claude4,claude45} has emerged as a critical phase in developing LLMs for programming tasks~\cite{fullstack,multiple,APPS_Benchmark,mceval,mdeval,Program_Synthesis_Code_LLM}, comprised of SFT and RL. SFT~\cite{wizardcoder,codearena,wavecoder,yang2025multiagentcollaborationmultilingualcode,zhang2025turningtiderepositorybasedcode,yang2024execrepobenchmultilevelexecutablecode} has become a cornerstone approach, where LLMs are fine-tuned on datasets of coding instructions paired with solutions to improve their ability to follow instructions. Further, RL-based approaches enable models to learn from environmental feedback and have gained strong performance across coding benchmarks, by incorporating feedback from code execution. Policy-based RL~\cite{grpo,dapo} methods are particularly effective for aligning models with diverse feedback signals in coding tasks.

\paragraph{LLM Factuality.}
LLM factuality~\cite{Locating_and_editing_factual_associations,Factool} is the ability of LLMs to produce factual content, including commonsense, world knowledge, and domain facts, substantiated by authoritative sources. Recent works~\cite{Locating_and_editing_factual_associations,Reward_collapse,LLM_Evaluators_Bias,LLM_Evaluators_SelfPreference,yang2025codesurvey,yang2025scalinglaws,chai2025multilingualmultimodalsoftwaredeveloper,zhang2025vgamegymvisualgamegeneration} explore LLMs as knowledge bases. To evaluate factuality, MMLU~\cite{mmlu} measures multitask accuracies, TruthfulQA~\cite{TruthfulQA} assesses the truthfulness of answers, and HaluEval~\cite{HaluEval} examines hallucinations. SimpleQA~\cite{simpleqa} and the search-augmented factuality evaluator (SAFE)~\cite{LongForm_Factuality} measure short-form and long-form factuality. Chinese SimpleQA~\cite{c_simpleqa} evaluates factuality in Chinese and factual-aware alignment~\cite{Factuality_Aware_Alignment,NewKnowledge_FT_Hallucinations,DawnAfterDark_Factuality} improves LLM factuality.

\section{Conclusion}

We propose \benchmark{}, a bilingual benchmark with 1,498 question-answer pairs across 15+ PLs and 21 domains, and \instruct{} with 66.9M training samples. Code factuality remains challenging even for frontier models, while \sftmodel{} and \rlmodel{} shows substantial improvements. RAG excels in frequently-updated documentation, while SFT handles stable concepts better, thinking mode outperforms chat mode logarithmically, and test-time scaling plateaus after initial gains. Results highlight the necessity of factuality-aware alignment beyond traditional execution metrics.

\clearpage
\section*{Limitations}
While CodeSimpleQA represents a significant contribution to evaluating factual code knowledge in LLMs, several limitations warrant acknowledgment. First, the benchmark's evaluation set comprises 1,498 carefully curated samples, which may not capture all edge cases or emerging technologies in the vast programming knowledge domain. Second, the benchmark currently supports only English and Chinese languages, limiting its applicability to other linguistic contexts. Third, our focus on short-form factual QA means we do not assess complex reasoning, multi-step problem-solving, or code generation quality. Fourth, the evaluation methodology relies on LLM-as-a-Judge, which may inherit biases from the judge model and could favor certain response styles. Fifth, the rapidly evolving nature of programming languages (PLs) and frameworks means that certain questions may become outdated, and emerging technologies may not be adequately represented. Finally, the human annotation process, though rigorous, involved a limited number of annotators, which may introduce subtle biases in question formulation and answer verification.

\section*{Ethics Statement}
The construction and release of CodeSimpleQA adheres to rigorous ethical standards throughout all stages of development. All annotators were fairly compensated and provided informed consent with comprehensive training on annotation guidelines. Source materials were exclusively drawn from publicly available programming documentation and open educational resources with proper attribution, and no personally identifiable information, proprietary code, or confidential data was included. The primary motivation for this research is to enhance factual accuracy of code-related LLM responses, potentially reducing software bugs, security vulnerabilities, and incorrect programming practices. While we acknowledge dual-use implications of improving LLM coding capabilities, we believe the benefits of accurate technical knowledge dissemination substantially outweigh potential risks, particularly since our benchmark focuses on factual understanding rather than vulnerability exploitation. To promote transparency and reproducibility, we commit to releasing both the CodeSimpleQA benchmark and construction methodology openly to the research community for responsible use in research, education, and improving AI system reliability.

\bibliography{custom}
\bibliographystyle{acl_natbib}

\clearpage
\appendix
\onecolumn

\section{Human Annotation}\label{appendix: Human Annotation}
To construct \benchmark{} as an evaluation benchmark covering a wide range of computer science topics, we implement a systematic human curation process involving 8 part-time annotators with expertise in multiple programming languages for data collection and annotation, and 3 senior software engineers for quality verification. All annotators participate in a curation tutorial and learn the annotation guidelines. The data collection process begins with manually gathering knowledge-bearing statements from prominent code documentation websites in both Chinese and English, including Stack Overflow, GitHub documentation repositories, official language documentation (e.g., Python.org, MDN Web Docs), and Chinese developer communities (e.g., CSDN, SegmentFault). Annotators then manually rewrite these factual code documentation statements into question-answer pairs, ensuring that questions test specific programming knowledge while answers remain grounded in the original documentation. Following best practices in dataset construction, we collect 1.5K initial samples and assign them to annotators for Question and Answer pair generation. The annotators successfully transform 680 suitable documentation statements into well-formed question-answer pairs. The process includes regular quality checks and feedback sessions to maintain consistency and accuracy throughout the annotation phase. The four senior software engineers independently review each QA pair to verify factual correctness, clarity, and alignment with the original documentation. Finally, 312 samples are retained (requiring consensus from at least 3 reviewers) to form CodeSimpleQA, resulting in a high-quality, bilingual dataset suitable for evaluating language models' factual knowledge of programming concepts and documentation.

\section{Prompts for Generating Chinese and English Factual QA Pair}
\begin{figure*}[h]
\begin{center}
    \includegraphics[width=1.0\textwidth]{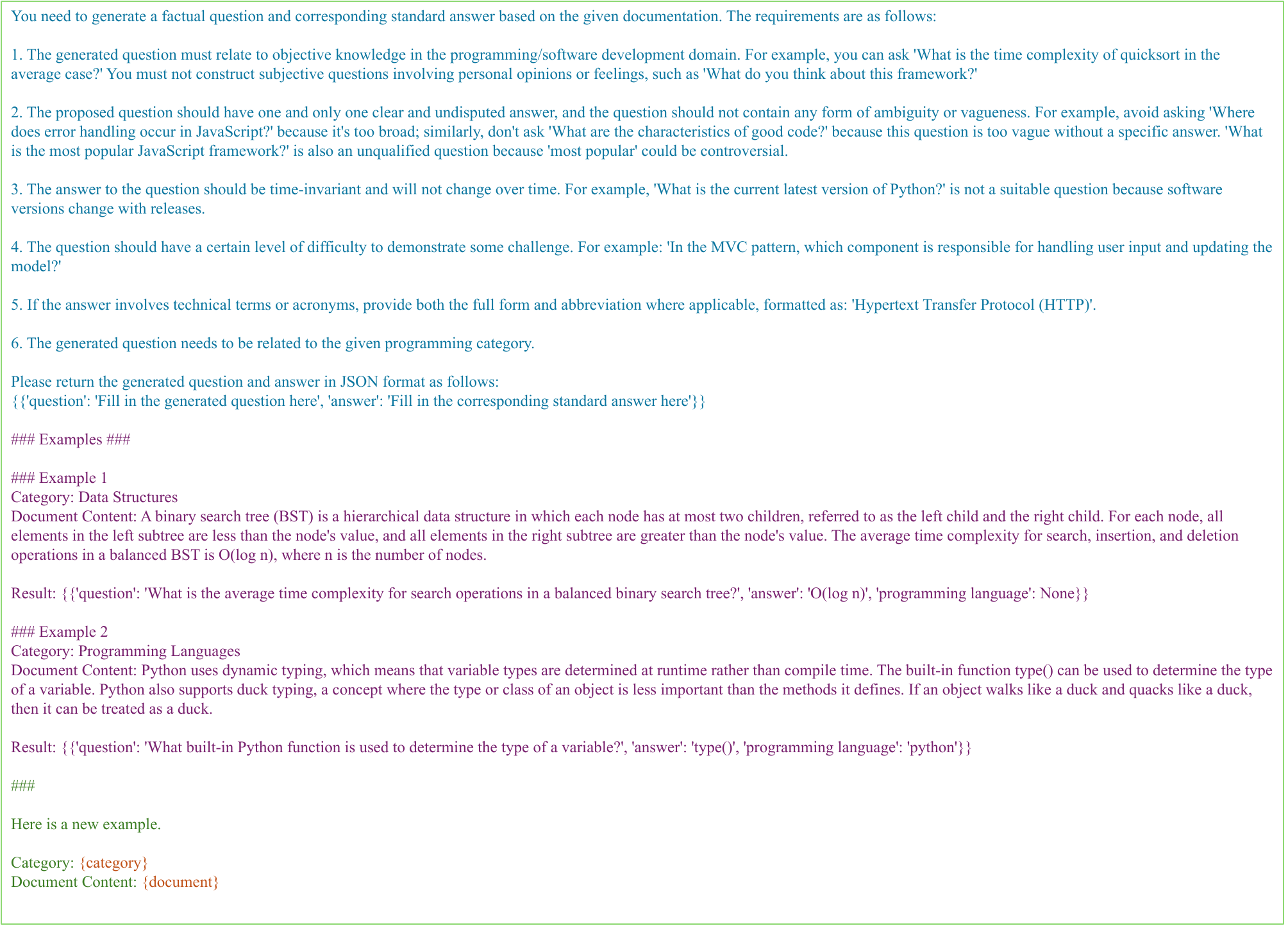}
    \caption{The prompts for generating English factual QA pairs.}
    \label{fig:evol_question_en}
\end{center}
\end{figure*}

\begin{figure*}[h]
\begin{center}
    \includegraphics[width=1.0\textwidth]{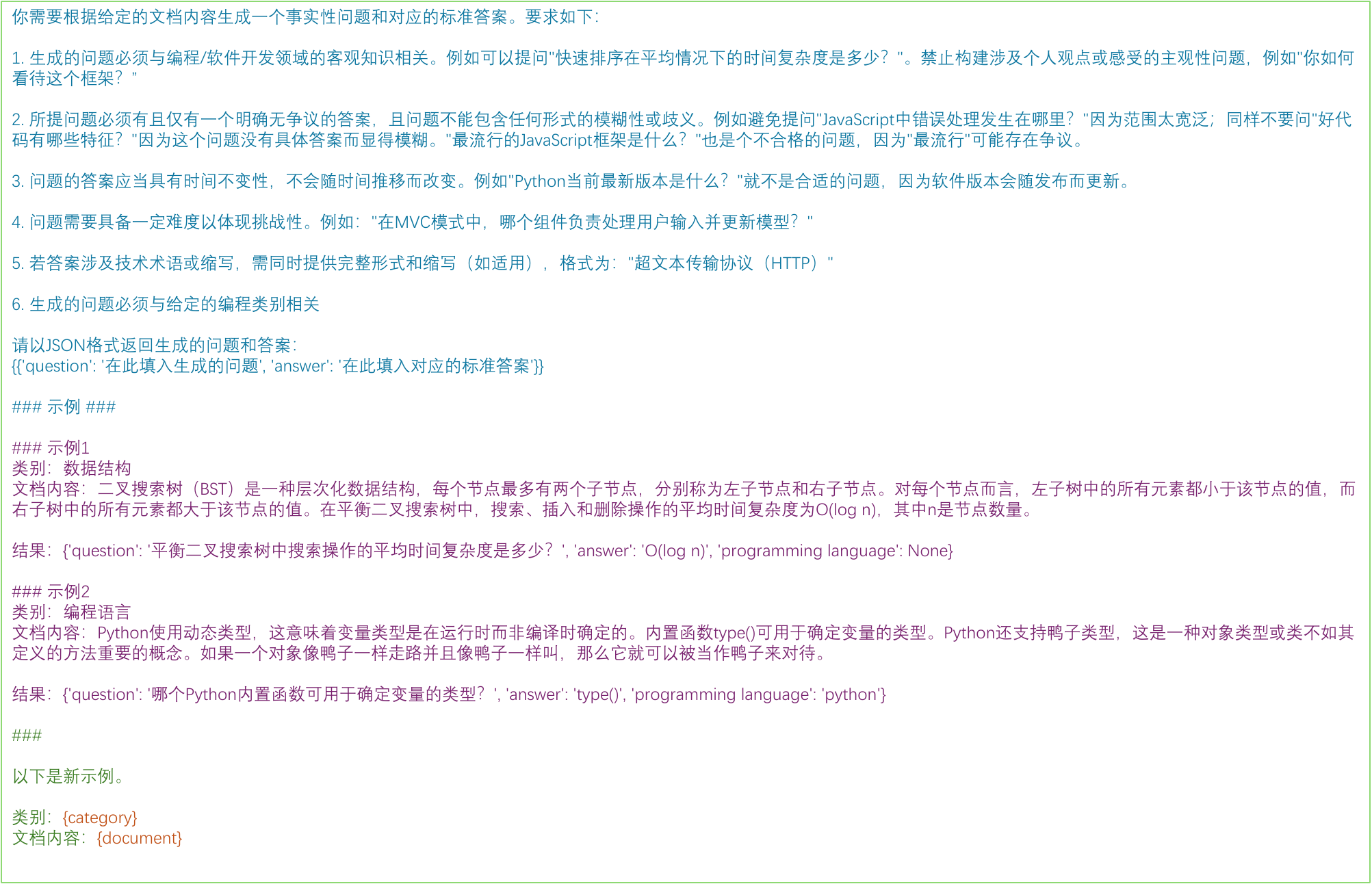}
    \caption{The prompts for generating Chinese factual QA pairs.}
    \label{fig:evol_question_zh}
\end{center}
\end{figure*}

\autoref{fig:evol_question_en} and \autoref{fig:evol_question_zh} are used to separately generate English and Chinese QA pairs. It provides guidelines for generating objective, unambiguous, time-invariant programming questions with single correct answers from documentation. The prompt includes detailed examples for each category and specific guidelines for code syntax, conceptual understanding, and numerical accuracy.

\end{CJK*}
\end{document}